\documentclass[journal,compsoc]{IEEEtran}
\ifCLASSOPTIONcompsoc
 \usepackage[nocompress]{cite}
\else
  \usepackage{cite}
\fi
\ifCLASSINFOpdf
\else
\fi

\usepackage[tight,normalsize,sf,SF]{subfigure}
\usepackage{color}

\usepackage{graphicx}
\usepackage[intlimits]{amsmath}
\usepackage{setspace}
\usepackage{multirow}
\usepackage{mwe}
\usepackage{amsfonts}
\usepackage{amssymb}
\usepackage{stmaryrd}
\usepackage{color}
\usepackage{diagbox}
\usepackage{booktabs}

\usepackage{hyperref}
\hypersetup{colorlinks,allcolors=black}

\newtheorem{definition}{Definition}

\usepackage{mathtools}

\begin{document}
%
\title{Spatiotemporal Propagation Learning for Network-Wide Flight Delay Prediction}
%
%
%
%

\author{Yuankai Wu \thanks{The authors are with the Department of Computer Science,
Sichuan University, Chengdu 610065, China (Corresponding author: Yuankai Wu. e-mail: wuyk0@scu.edu.cn)}, Hongyu Yang, Yi Lin, Hong Liu}
\IEEEtitleabstractindextext{%
\begin{abstract}
Demystifying the delay propagation mechanisms among multiple airports is fundamental to precise and interpretable delay prediction, which is crucial during decision-making for all aviation industry stakeholders. The principal challenge lies in effectively leveraging the spatiotemporal dependencies and exogenous factors related to the delay propagation. However, previous works only consider limited spatiotemporal patterns with few factors. To promote more comprehensive propagation modeling for delay prediction, we propose SpatioTemporal Propagation Network (STPN), a space-time separable graph convolutional network, which is novel in spatiotemporal dependency capturing. From the aspect of spatial relation modeling, we propose a multi-graph convolution model considering both geographic proximity and airline schedule. From the aspect of temporal dependency capturing, we propose a multi-head self-attentional mechanism that can be learned end-to-end and explicitly reason multiple kinds of temporal dependency of delay time series. We show that the joint spatial and temporal learning models yield a sum of the Kronecker product, which factors the spatiotemporal dependence into the sum of several spatial and temporal adjacency matrices. By this means, STPN allows cross-talk of spatial and temporal factors for modeling delay propagation. Furthermore, a squeeze and excitation module is added to each layer of STPN to boost meaningful spatiotemporal features. To this end, we apply STPN to multi-step ahead arrival and departure delay prediction in large-scale airport networks. To validate the effectiveness of our model, we experiment with two real-world delay datasets, including U.S and China flight delays; and we show that STPN outperforms state-of-the-art methods. In addition, counterfactuals produced by STPN show that it learns explainable delay propagation patterns. The code of STPN is available at \url{https://github.com/Kaimaoge/STPN},

\end{abstract}

}

\maketitle

\section{Introduction}

Flight delay is a significant problem faced by the modern aviation industry. In 2019, it was estimated that the global economy's annual cost of flight delays is approximately \$ 50 billion \cite{IATA}. Unfortunately, most delay mitigation measures are costly, challenging to implement, or both. For instance, constructing new airports to ease delays is difficult due to the high cost. In the U.S., the average cost to build a commercial airport is \$30 million per 3 km runaway and \$500 per square meter for an airport passenger terminal \cite{airport_cost}. More seriously, the aviation system allows for little room to accommodate deviations. As a result, the system is slow to respond to unexpected events leading to potentially minor local delays that cascade into network-wide congestion. With the advancements and widespread adoption of information technology, access to large flight databases is now available. It has led to the development of flight delay prediction as a research field. Advanced air traffic management \cite{bertsimas2011integer, bertsimas2016fairness, ayhan2016aircraft} made through short-term and long-term predictions is a far cheaper and more accessible alternative for reducing flight delays. Moreover, the predicted information is also beneficial for adjusting travelers' pre-trip schedules and reducing unnecessary anxiety. 

\begin{figure}[!ht]
    \centering
    \includegraphics[width = 0.48\textwidth]{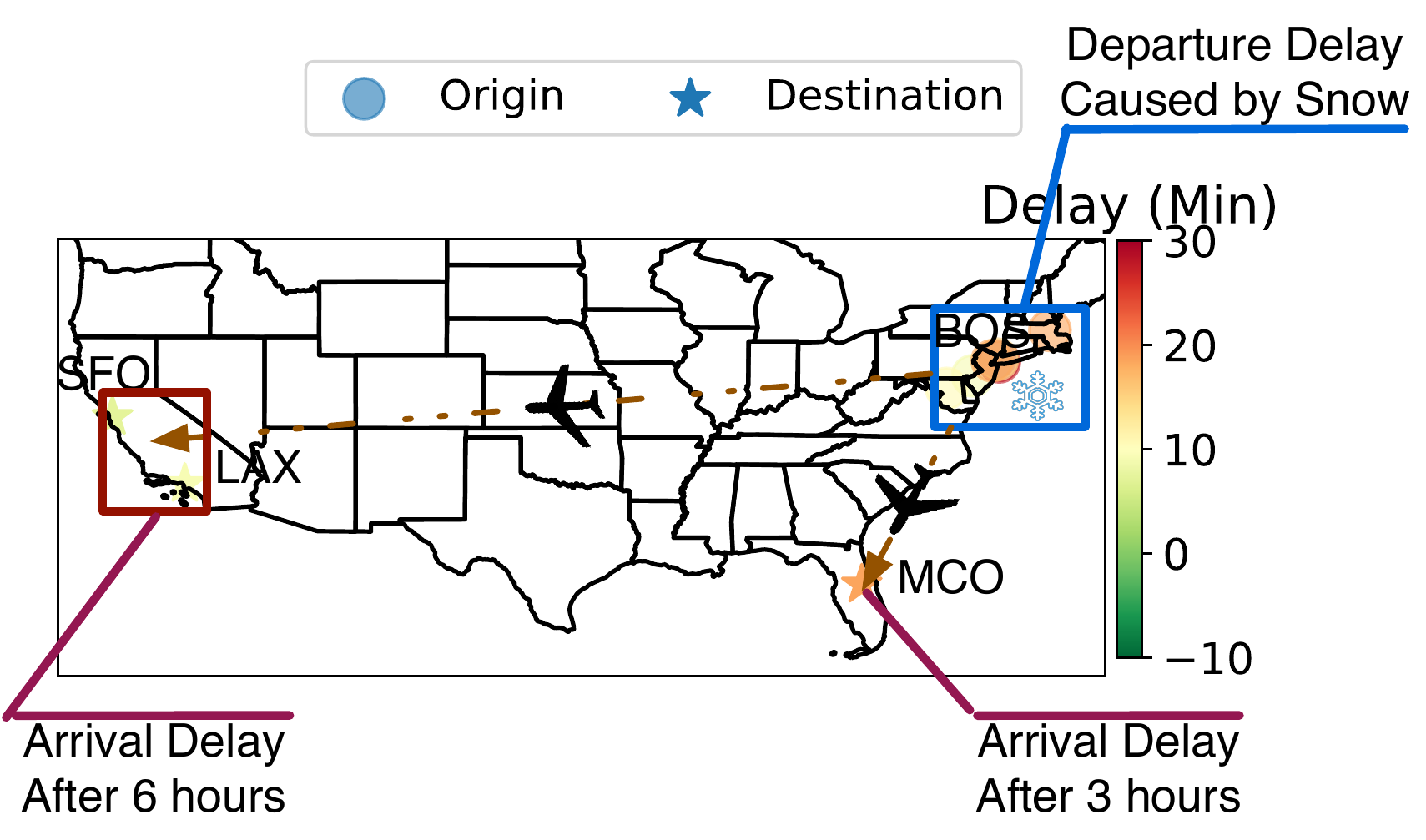}
    \caption{Spatiotemporal dependencies within the airport network.}
    \label{fig:delay_prop}
\end{figure}

Flight delay prediction is different from conventional time-series analysis in that delay propagation is subject to spatial and many other external factors. For instance, the prediction of delay at one airport depends on the delay at other related airports \cite{kafle2016modeling}. In addition, all of the airports are affected by external factors such as extreme weather. In other words, how to learn the spatiotemporal dependencies within an airport network is essential for large-scale delay prediction. 
Unfortunately, the spatiotemporal dependencies are very complex due to various factors such as geographic proximity, weather condition, and airline schedule. Figure~\ref{fig:delay_prop} illustrates an example. Powerful snow causes flight departure delays in east coast airports, including Boston Logan International (BOS), New York LaGuardia (LGA), and John F. Kennedy International (JFK) airports. Due to their geographic proximity, the departure delays at these airports are positively correlated under adverse weather conditions. The aircraft is delayed in departing from these airports and will almost certainly arrive late at its destination airports. As a result, these departure delays lead to long-range arrival delay propagation in Orlando International (MCO), Los Angeles International (LAX), and San Francisco International (SFO) airports due to their high traffic volume from the east coast airports. Moreover, we can observe that the delay level of LAX and SFO are weaker than that of MCO, partially because they are comparatively far away from east coast airports. This example shows that the complexity of spatiotemporal dependencies within airport network delay lies in the following four aspects:

\textbf{1.Exogenous Factors:} The operational characteristics of air transportation render it vulnerable to various exogenous factors, e.g., extreme weather conditions like high winds, low visibility, and thunderstorms are the primary cause of air flight delays, accounting for almost 40\% of flight delays \cite{li2021graph}. 

\textbf{2.Multi-relational Spatial Dependencies:}  The delays within the network can show strong local similarities between nearby airports due to the exogenous factors only affecting a small regional area. However, the delays also exhibit long-range spatial dependencies because the delayed flight can propagate delays between two airports far from each other.

\textbf{3.Coupled Spatiotemporal Effects:} The delay propagation in an airport network is transferred and amplified by flights connecting different airports. As a result, the delay propagation between two airports depends on the flight's travel time and the distance between airports. This operation characteristic makes the spatial and temporal dependencies coupled with each other \cite{cai2021spatial}.  

\textbf{4. Departure-Arrival Delay Relationship:} The departure and arrival delays are most likely simultaneously happen on one flight. If the delay time is long, several related flights will be departing and arriving delayed in several different airports.

Given the aforementioned complexity of airport network delay, delay prediction has been a hot topic for decades, falling into two main categories: knowledge-driven and data-driven approaches. In transportation and operational research, previous works have focused on modeling the dynamics of delay propagation using queuing theory \cite{pyrgiotis2013modelling, simaiakis2016queuing}. The knowledge-driven approaches are explainable and help identify factors that mitigate or amplify delay propagation. However, flight delay data is routinely produced in high-volume and high-dimension and can not be easily handled via classical knowledge-driven approaches. Several works have applied machine learning algorithms including random forest, Gradient Boosting Regression Tree (GBRT), and K-nearest neighbor algorithm \cite{rebollo2014characterization} for the single airport and network-wide delay prediction. A problem with traditional machine learning methods is that their shallow structure can not efficiently handle spatial dependencies within big data. Most recently, deep learning models for delay prediction have been developed \cite{yu2019flight}, but without considering the spatial dependencies. In \cite{bao2021graph} and \cite{cai2021deep}, the spatial dependencies are modelled by Graph Convolutional Networks (GCNs). However, those approaches only considered limited factors using a simple structure without incorporating the aforementioned characteristics including the multi-relational spatial dependencies, coupled spatiotemporal effects, and departure-arrival delay relationships.

In this paper, we study how to simultaneously predict multi-step ahead arrival and departure delay in a large-scale airport network. To fully characterize the complex spatiotemporal dependencies within the airport network, we represent the airport network using a multi-relational graph whose nodes are airports. The multi-relation between nodes is characterized by different edge weights measured by airport distances and flight volume. We model the delay propagation process within the constructed multi-relational graph using a Space-Time-Separable Graph Convolutional Network (STSGCN) \cite{sofianos2021space}. We further propose the Spatiotemporal Propagation Network (STPN) that integrates the self-attention mechanism \cite{vaswani2017attention} for learning temporal dependencies, diffusion convolution \cite{li2018diffusion} for capturing spatial dependencies, and squeeze-and-excitation \cite{hu2018squeeze} for modeling feature relationships.  When evaluated on real-world delay datasets, STPN consistently outperforms state-of-the-art traffic forecasting baselines. In summary:
\begin{itemize}
\item We propose our model under the space-time separable graph convolution network scheme. Unlike the existing model, we use multiple graph structures in time and space modes. Theoretical analysis through tensor algebra shows that our model learns a sum of Kronecker products kernel, rather than a simple Kronecker kernel  \cite{sofianos2021space}. It makes our model can learn more complex spatiotemporal dependencies. 
\item We model the multiple spatial dependencies via a random walk process on a multi-relational graph. We use diffusion graph convolution to characterize the random walk process.
\item The multi-head self-attention mechanism is introduced to model the temporal dependencies. Multiple temporal adjacency matrices learned from multi-head self-attention can simultaneously capture various kinds of temporal dependencies. 
\item Extensive experiments conducted on two large-scale
datasets show the effectiveness of the proposed method
for both arrival and departure delay prediction. We also use our model to produce counterfactuals, which show our model learns explainable delay propagation patterns.
\end{itemize}

The rest of this paper is organized as follows. First, we review some related works of flight delay propagation modeling and graph neural networks for spatiotemporal prediction in Section~\ref{sec:rel}.  We then introduce the proposed approach for network-level flight delay prediction in Section~\ref{sec:meh}. Extensive comparison, ablation analysis and counterfactual intervention are conducted in Section~\ref{sec:exp}. Finally, we conclude this paper in Section~\ref{sec:con}.

\section{Related Works}
\label{sec:rel}
\subsection{Network-wide delay propagation modeling}

The earliest study on flight delay propagation can date back to 1998 \cite{beatty1999preliminary}. The propagated delay occurs because of connected resources involved in an initially delayed flight and flights downstream. Using the flight schedule, Beatty et al. \cite{beatty1999preliminary} construct a delay tree containing 50-75 connecting flights. Later, this simple approach was extended to study network-wide airport congestion \cite{fleurquin2013systemic}. The most successful analytical model for studying delay propagation is the Approximate Network Delays model (AND) \cite{pyrgiotis2013modelling}, which employs a combination of a queuing model for simulating initial delays and a delay-propagation algorithm. The AND model mainly uses flight schedules to simulate delay propagation and can uncover rich temporal dependencies of airport delays. Then, several applications \cite{campanelli2016comparing}, and improvements \cite{kafle2016modeling, wu2019modelling} under the umbrella of AND are proposed. Other analytical models like multivariate simultaneous equation regression show that major airports have a higher impact on delay propagation \cite{nayak2011estimation}. 

A recent study \cite{li2021graph} suggests that graph signal processing is a promising tool for studying delay propagation. For example, the airport delay can be treated as node signals in a graph, and various spatiotemporal patterns can be qualified by graph spectral analysis. Some attempts also utilize GCNs to learn delay propagation. Bao et al. \cite{bao2021graph} propose AG2S-Net, which models the spatial dependency as learning parameters. Cai et al. \cite{cai2021deep} propose MSTAGCN, which uses feature embeddings to weight the spatial interactions between different airports. A drawback of those two models is that they treat spatial and temporal dependencies separately, while those two are bonded. Compared with those two approaches, our STPN models the spatial dependency more systematically, i.e., generalizing convolution to the airport network graph based on the nature of delay propagation. Besides, we derive multi-graph convolution from the property of delay propagation related to geographic proximity, time, and flight schedule factors. Specifically, the time factors on delay propagation are modeled by the Space-Time-Separable graph convolution scheme, where a special-designed self-attention mechanism learns the temporal adjacency matrices.


\subsection{Graph neural networks for spatiotemporal prediction}

GNNs are divided into two main categories, the spectral-based approaches and spatial-based approaches~\cite{wu2019comprehensive}. Most existing spatiotemporal prediction frameworks are based on the spectral-based Graph Convolutional Network (GCN), which is initially proposed by Bruna et al. \cite{bruna2014spectral}. Since then, several ideas \cite{kipf2016semi, defferrard2016convolutional, atwood2016diffusion, zhu2020simple} have been proposed to improve the performance of GCN. 

In the context of spatiotemporal prediction, the GNN-based approaches treat sensors or locations as nodes of a graph, and establish edges according to their spatial relationships. Then the spectral-based GCNs are utilized to capture the spatial dependencies of the established graph signals. To simultaneously model the temporal dependencies, GCNs are combined with recurrent neural networks (RNNs), temporal convolutional networks (TCNs), and self-attentional mechanism. Seo et al. \cite{seo2018structured} used GCNs to filter inputs and hidden states in RNNs. Later, Li et al. \cite{li2018diffusion} combined RNNs with diffusion convolution for long-term traffic forecasting, in which the effects of asymmetric spatial dependencies can be taken into account. To reduce the computational cost brought by RNNs, Yu el al. \cite{yu2018spatio} built a complete convolutional structure named STGCN, in which a specifically designed TCN is used to capture the temporal dependencies. The earlier spatiotemporal prediction approaches were limited to graphs constructed by geographic proximity and incapable of uncovering other types of spatial dependencies. Wu et al. \cite{wu2019graph} mitigated this issue using a self-adaptive adjacency matrix. As a result, this enabled capturing spatial dependencies from farther sensors, but it did not consider dynamic spatial dependencies. To address this drawback, Zhang et al. \cite{zheng2020gman} introduced the GMAN model that dynamically assigns different attention weights to different sensors at different time steps. Guo et al. \cite{guo2021learning} further combined dynamic graph convolution with Transformer to capture the temporal dynamics of spatiotemporal data. As opposed to previous efforts, Sofianos et al. \cite{sofianos2021space} used a pure GCN structure to handle spatial and temporal dependencies. They factorized spatiotemporal graphs into space and time adjacency matrices, which are end-to-end learnable. Their method achieves state-of-the-art performance on human pose forecasting and gives us a lot of inspiration. Another related work is the attempt \cite{monti2017geometric} to generalize multiple graph convolution on matrix completion tasks, in which two GCNs are applied to each dimension of the matrix.

\section{Methodology}
\label{sec:meh}
\begin{figure*}[!ht]
    \centering
    \includegraphics[width = 0.8\textwidth]{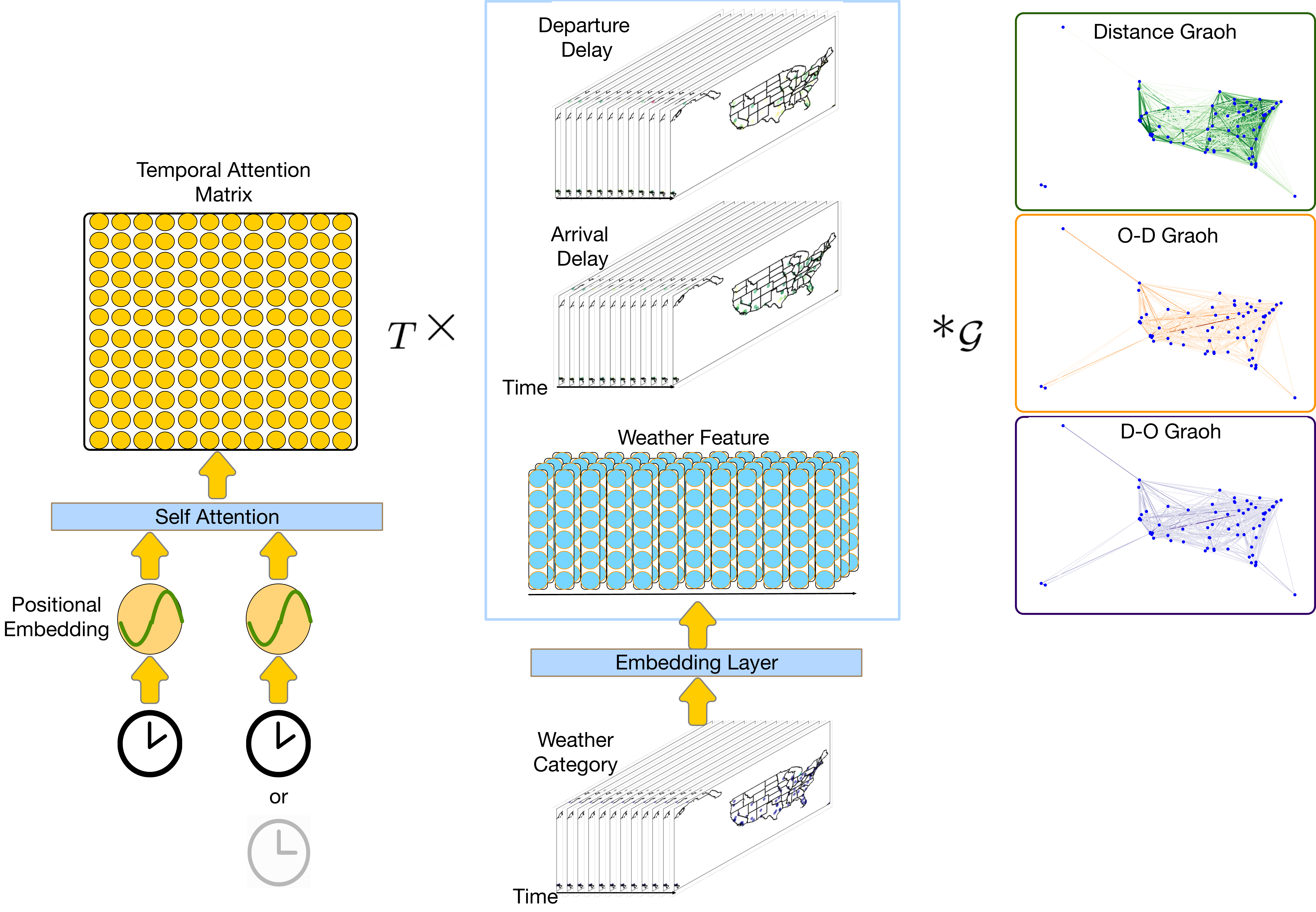}
    \caption{Overview of the basic building block (Input layer) of STPN. Given a sequence of historical departure/arrival delays and weather conditions of an airport network, STPN captures temporal dependencies by self-attention and allows diverse spatial propagation through geographic proximity and origin-destination traffic volume.}
    \label{fig:block}
\end{figure*}

The proposed model leverages the spatiotemporal propagation patterns learned from historical departure/arrival delays and external factors (weather conditions) to forecast the long-term future departure/arrival delays. The propagation patterns are learned by space-time separable multi-graph convolution, which considers the joint space-time interaction between airports. Figure~\ref{fig:block} illustrates the basic building block of STPN. The effects of geographic proximity, weather conditions, and traffic volume on delay propagation are incorporated into STPN. In this section, we further provide insights into the STPN model.

\subsection{Problem Formalization}

The airport network delay prediction problem in this paper is the following. Given a set of $N$ airports, we represent those airports as a weighted multi-relational graph $\mathcal{G} = \{ \mathcal{V}, \mathcal{E}, \mathcal{R} \}$, where $\mathcal{V}$ is a set of nodes $|\mathcal{V}| = N$, $\mathcal{R}$ is a set of $Q$ relations, and $\mathcal{E} \subseteq \mathcal{V} \times \mathcal{V} \times \mathcal{R}$ is a set of $m$ weighted edges. Given $Q$ relations, we can define $Q$ adjacency matrices $[A_1, \cdots, A_Q]$ with $A_q \in \mathrm{R}^{N \times N}$. We denote the arrival and departure delays by 2-dimensional vectors $\mathbf{x}_{v, k}$ representing delays of airport $v$ at time $k$, the covariate (etc. air condition) vectors $\mathbf{z}_{v, k}$ representing weather type of airport $v$ at time $k$. Denote the delays observed on graph $\mathcal{G}$ as a matrix $X \in \mathrm{R}^{N \times 2}$, and the covariates as a matrix $Z \in \mathrm{R}^{N \times C}$ with $C$ categories. The delay historical observation from time point $t - h + 1$ to $t$ is denoted by a  3-D tensor $\mathfrak{X}_{t-h+1: t-1} = [X_{t-h+1}, \cdots, X_t] \in \mathrm{R}^{N \times h \times 2}$, and historical covariates are denoted by a 3-D tensor $\mathfrak{Z}_{t-h+1: t-1} = [Z_{t-h+1}, \cdots, Z_t] \in \mathrm{R}^{N \times h \times 2}$. The aim is to learn a function $f(\cdot)$ that maps $h$ historical observations and covariates to future $p$ delays, given a multi-relational graph $\mathcal{G}$:
\begin{equation}
   (\mathfrak{X}_{t-h+1: t-1}, \mathfrak{Z}_{t-h+1: t-1}, \mathcal{G}) \overset{f(\cdot)}{\to} [X_{t}, \cdots, X_{t+p}].
\end{equation}

\subsection{Space-time Separable Graph Convolution}

The STPN model is parital inspired by STSGCN~\cite{sofianos2021space}. To make the proposed STPN more understandable, we use tensor algebra to emphasize STSGCN. The input of the traditional graph convolution layer $l$ is a matrix $H^{(l)} \in \mathrm{R}^{N \times C^{(l)}}$, where $N$ is the number of nodes, and $C^{(l)}$ is the number of features. The graph convolution layer $l$ outputs the $H^{l+1} \in \mathrm{R}^{N \times C^{l+1}}$, given by the following:
\begin{equation}
    H^{l+1} = \sigma(A H^{(l)} W^{(l)}),
\end{equation}
where $A \in \mathrm{R}^{N \times N}$ is the normalized adjacency matrix, $W^{(l)} \in \mathrm{R}^{C^{(l)} \times C^{l+1}}$ are the trainable
weights of layer $l$,  and $\sigma$ is the activation function.

Different from the basic graph convolution layer, the spatiotemporal input of the STSGCN layer is a tensor $\mathcal{H}^{(l)} \in \mathrm{R}^{N \times T^{(l)} \times C^{(l)}}$. We define STSGCN via tensor mode product\cite{kolda2009tensor}:
\begin{definition}
  The $n$-mode (matrix) product of a tensor $\mathcal{X} \in \mathrm{R}^{I_1 \times \cdots \times I_N}$ with a matrix $U \in \mathrm{R}^{I_n \times J}$ is denoted by $\mathcal{X} \times_n U$ is of size $I_1 \times \cdots \times I_n \times J \times \cdots \times I_N$. Elementwise, we have
  \begin{equation*}
      (\mathcal{X} \times_n U)_{i_1\cdots i_{n-1}j i_{n+1} \cdots i_N} = \sum^{I_n}_{i_n =1} x_{i_1 i_2 \cdots i_N} u_{j i_n}.
  \end{equation*}
\end{definition}
For input tensor $\mathcal{H}^{(l)}$, we have space ($S$), time ($T$), and feature ($F$) modes. The graph convolution of STSGCN can be defined as follows:
\begin{equation}
    \mathcal{H}^{l+1} = \sigma \left ({\mathcal{H}^{(l)} \times_S A^{(l)}_S \times_T A^{(l)}_T \times_F W^{(l)}}\right ),
    \label{eq:tensor_graph}
\end{equation}
where $A^{(l)}_S \in \mathrm{R}^{N \times N}$ and $A^{(l)}_T \in \mathrm{R}^{T^{(l)} \times T^{(l+1)}}$ are spatial and temporal adjacency matrices, respectively, and $\mathcal{H}^{(l+1)} \in \mathrm{R}^{N \times T^{(l+1)} \times C^{(l+1)}}$ is the output. The tensor form given in Equation~\eqref{eq:tensor_graph} is equivalent to the following matrix form:
\begin{equation}
    H^{l+1} = \sigma \left( \left (  A^{(l)}_S \otimes A^{(l)}_T \right )^\intercal H^{(l)} W^{(l)} \right),
\end{equation}
where $H^{(l)} \in \mathrm{R}^{NT^{(l)} \times NT^{(l)}}$, $\otimes$ is the Kronecker product combining $A^{(l)}_S$ and $A^{(l)}_T$ into an $NT^{(l)} \times NT^{(l+1)}$ matrix block. $A^{(l)}_S \otimes A^{(l)}_T$ is the Kronecker product kernel, widely used in Gaussian Processes (GPs) \cite{bonilla2007multi}. It reduces the model parameters and avoids inference of full spatiotemporal kernel. However, the Kronecker product kernel assumes the spatial dependencies at best change by a factor in time, hence is unable to characterize more complex spatiotemporal dependencies \cite{lan2019learning}.

The original STSGCN \cite{sofianos2021space} treats $A^{(l)}_S$ and $A^{(l)}_T$ as purely trainable weights for neural networks. It will inevitably add learning parameters to the neural networks, and thus is not feasible for a graph with a large number of nodes. Moreover, learning $A^{(l)}_S$ and $A^{(l)}_T$ with fixed sizes will make the neural networks transductive, and full retraining is required when we have a new node in the graph (a newly constructed airport in the system) \cite{wu2021inductive}. Another drawback is that the purely learned adjacency matrix neglects the prior knowledge and external factors related to the spatiotemporal dependencies. To enhance the generalization capability of STPN, we use an inductive architecture to infer spatiotemporal dependencies, making full use of prior knowledge about the aviation system. The following sections will introduce the 
spatiotemporal dependencies learning methods of the proposed STPN.

\subsection{Self-attention for Temporal Mode}

The self-attention layer for temporal mode is to obtain a representative temporal adjacency matrix $A^{(l)}_T$. To account for the daily periodicity and rhythms of flight schedule, we introduce positional encoding \cite{vaswani2017attention} for the time of day:
\begin{equation}
\begin{split}
\mathbf{pe}(pos_n, 2i) &= \sin\left(pos_n/L_{pos}^{{2i}/{J}}\right), \\
\mathbf{pe}(pos_n, 2i+1) &= \cos \left(pos_n/L_{pos}^{{2i}/{J}}\right),
\end{split}
\end{equation}
where $pos_n\in \{0, 1,\cdots,t_d\}$ is the time of day, $t_d$ is the maximum value of daily time determined by the temporal resolution of delay data, $J$ is the total dimension of the embedding, $i \in \{1, \ldots, \lfloor J/2 \rfloor \}$, and $L_{pos}$ is the scaling factor. Although a fixed $L_{pos}$ (e.g, 10,000) is often selected in defining positional embedding, we treat it as a learning parameter because it could be beneficial for general approximation \cite{wang2020position}. 

Given a series of historical delays $\mathfrak{X}_{t-h+1: t-1}$ and their associated embedding $P_{t-h+1: t-1} = [pe_{t-h+1}, \cdots, pe_{t-1}] \in \mathrm{R}^{h \times J}$. We compute the temporal adjacency matrix $A^{(l)}_T$ by employing the self-attention mechanism. The multi-head self-attention layer transforms the embedding $P_{t-h+1: t-1}$ into query matrix $Q^{(l)} = P_{t-h+1: t-1} W^{(l)}_Q$ and key matrices $K^{(l)} = P_{t-h+1: t-1} W^{(l)}_K$, Here, $W^{(l)}_Q, W^{(l)}_K \in \mathrm{R}^{J \times c_k}$ are learnable parameters of the $l$-th layer. After these linear projections, the scaled dot-product attention computes the adjacency matrix:
\begin{equation}
    A^{(l)}_T = {\operatorname{softmax}}\left ( {\frac {Q^{(l)}(K^{(l)})^{\intercal  }}{\sqrt {c_{k}}}} \right )^\intercal.
    \label{eq:att}
\end{equation}
We apply self-attention to the hidden layer of STPN. Since the attention mechanism is order-independent, we can provide positions for any time of day and compute the associated adjacency matrix. For the output layer, the target is to obtain $[X_{t}, \cdots, X_{t+p}]$, and we have positional embedding $P_{t: t+p} = [pe_{t}, \cdots, pe_{t+p}] \in \mathrm{R}^{p \times J}$. The query for the output layer will be $Q^o = P_{t: t+p} W^{(l)}_Q$, then we have $A^{o}_T \in \mathrm{R}^{h \times p}$. Using tensor mode product $\mathcal{H}^o \times_T A^{o}_T$, the temporal size of the outputs will change from $h$ to $p$. 

We can also generate multiple temporal adjacency matrices using Equation~\eqref{eq:att} to capture multiple temporal dependencies better, it will yield multiple temporal representations $[\mathcal{H} \times_T A_{T,1}, \cdots, \mathcal{H} \times_T A_{T,I}]$ (multi-head attention). We will use those representations jointly with multiple graphs of spatial mode for modeling spatiotemporal dependencies.

\subsection{Multi-Graph Convolution for Spatial Mode}

We consider that the spatial dependencies of flight delay arise from a multi-relational graph and relate the spatial propagation of delay to random walks. Let the vector $\mathbf{p_t} \in \mathrm{R}^N$ denote the delay probability distribution on a multi-relational graph,  $\mathbf{p_t}(n)$ indicate the probability of being at node $n$ at time $t$ ($\sum^N_n \mathbf{p_t}(n) = 1$), $w_1, \cdots, w_Q$ denote the probability of taking random walk according to relation $1, \cdots, Q$ with ($\sum^Q_q w_q = 1$). To derive $\mathbf{p_{t+1}}$ from $\mathbf{p_{t}}$, the random walk process can be stated as:
\begin{equation}
   \mathbf{p_{t+1}} = \sum^Q_{q=1} w_q \hat{A}_q \mathbf{p_{t}},  
\end{equation}
where $\hat{A}_q$ 
represents the power series of the transition matrix with $\hat{A}_q = {A}_q/\text{rowsum}(A_q)$. This process can be modelled by a diffusion convolution layer,
which proves to be effective in spatial-temporal modeling \cite{li2018diffusion,wu2019graph}. The diffusion convolution layer can be generalized to the following equation:
\begin{equation}
    X \ast_{\mathcal{G}} = \sum^K_k \sum^Q_q \hat{A}^k_q X W_{q,k},
    \label{eq:random_walk}
\end{equation}
where $K$ is the number of diffusion steps, $W_{q,k}$ are learnable weights. Using multiple temporal representations jointly with diffusion convolution, we yield the following tensor algebra form:
\begin{equation}
    \mathcal{H}^{l+1} =  \sigma \left (\sum^K_k \sum^Q_q \sum^I_i {\mathcal{H}^{(l)} \times_S \hat{A}^k_q \times_T A^{(l)}_{T, i} \times_F W^{(l)}_{q,k}} \right ).
    \label{eq:tensor_multiple_graph}
\end{equation}
Ignoring the learnable weights $W^{(l)}_{q,k}$ on feature mode, we have the following matrix form:
\begin{equation}
\sum^K_k \sum^Q_q \sum^I_i \left (\hat{A}^k_q  \otimes A^{(l)}_{T, i} \right )^\intercal {H}^{(l)}. 
\end{equation}
$\sum^K_k \sum^Q_q \sum^I_i \left (\hat{A}^k_q  \otimes A^{(l)}_{T, i} \right )^\intercal$ is equal to the sum of Kronecker products kernel in Gaussian Processes \cite{wilson2014fast}. In the sum of Kronecker products kernel, each term $\hat{A}^k_q  \otimes A^{(l)}_{T, i}$ presents a combination of one spatial and temporal dependencies. Unlike the single Kronecker products kernel used in STS-GCN \cite{sofianos2021space}, the sum model allows for multiple temporal evolutions with specific spatial patterns and can, thus, account for temporal nonstationarities in separate terms. Using the sum of Kronecker products kernel, we can capture the spatiotemporal propagation pattern of airport delays and account for multiple kinds of spatial dependencies.

\subsection{Squeeze-and-Excitation on Feature Mode}

\begin{figure}[!ht]
    \centering
    \includegraphics[width = 0.3\textwidth]{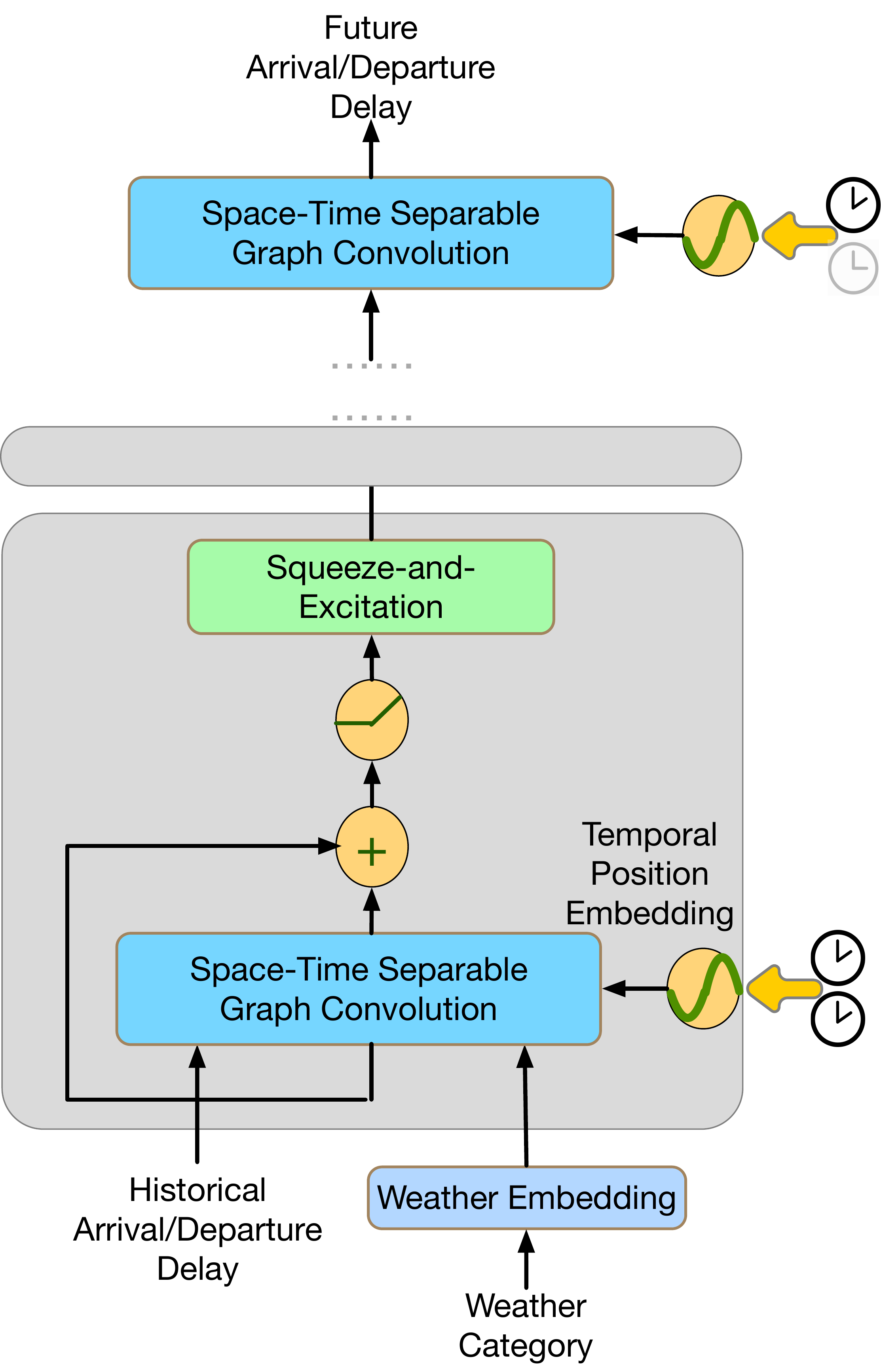}
    \caption{STPN architecture. STPN stacks multiple Space-Time-Separable Multi-Graph Convolution layers defined in Equation~\eqref{eq:tensor_multiple_graph} and Squeeze-and-Excitation layers. The output layer is also a Space-Time-Separable Multi-Graph Convolution layer with a future time point query vector.}
    \label{fig:structure}
\end{figure}

In our STPN, the arrival delays, departure delays, and embeddings of the weather category are directly treated as features of the graph neural networks. Those features are related to each other. In the hidden layer $l$, we have a feature map tensor $\mathcal{H}^{(l)}$ of size $N \times T^{(l)} \times C^{(l)}$ with $C^{(l)}$ feature maps. The relationship between different features are captured by fully connected layer with learnable weights $W^{(l)}_{q,k}$. We assume that these feature maps are redundant and have a different magnitude of importance for delay prediction. To make STPN more sensitive to informative features, we add a squeeze-and-excitation (SE) block \cite{hu2018squeeze} on feature mode. Given the feature map $\mathcal{H}^{(l)}$, we have squeeze vector $\mathbf{z}^{(l)} \in \mathrm{R}^{C^{(l)}}$ whose $c$-th value equals to
\begin{equation}
    z^{(l)}_c  = \frac{1}{N \times T^{(l)}} \sum^N_{i=1} \sum^{T^{(l)}}_{j=1} \mathcal{H}^{(l)}(i, j, c).
\end{equation}
Then we have a gating mechanism:
\begin{equation}
\mathbf{s}^{(l)} = \text{Sigmoid}\left (W^{(l)}_{SE1} \text{ReLU} \left (W^{(l)}_{SE2} \mathbf{z}^{(l)}\right ) \right ),
\end{equation}
where $W^{(l)}_{SE1} \in \mathrm{R}^{\frac{C^{(l)}} {r} \times C^{(l)}}$ and $W^{(l)}_{SE2} \in \mathrm{R}^{ C^{(l)} \times \frac{C^{(l)}} {r}}$. Finally, the $c$-th feature map is produced by 
\begin{equation}
    \hat{\mathcal{H}}^{(l)}(:, :, c) = s^{(l)}_c  {\mathcal{H}}^{(l)}(:, :, c).
\end{equation}
In our experiments, we find that the SE block can slightly improve our model's performance. To this end, our STPN consists of several space-time-separable multi-graph convolution layers given in Equation~\eqref{eq:tensor_multiple_graph} with residual connections PReLU activation followed by an SE block. The output layer is a space-time-separable multi-graph convolution layer with a future time point query vector and Linear activation. In summary, Figure~\ref{fig:structure} shows the overall architecture of STPN.

\subsection{Training}

The proposed architecture is trained end-to-end supervisedly. The model is trained by Root Mean Squared Error (RMSE):
\begin{equation}
    L_{RMSE} = \sqrt{\frac{1}{Npc} \sum^N_{n = 1} \sum^p_{i=0} \sum^2_{c= 1} \left ( X_{t+i}(n, c) - \bar{X}_{t+i}(n, c) \right )^2}. 
\end{equation}
It should be noted that some airports do not have flights for some periods. We treat those data points as missing data and mask their training losses.

\section{Experiments}
\label{sec:exp}

In this section, we assess the quality of delay prediction in two different datasets and perform ablation studies to isolate the impact of each building block in the proposed regime. 

\subsection{Evaluation Metrics}

Throughout all experiments, we predict both network-wide arrival and departure delay. This yields delay observation $\mathbf{x}_{v, k} \in \mathrm{R}^2$ at time point $k$ on airport $v$. We calculate the errors for arrival and departure delays separately.

To evaluate the quality of delay prediction, we use three common metrics: mean absolute error $MAE$, root mean squared error $RMSE$, R squared $R^2$. Letting $\mathrm{J}$ denote the prediction set, $\mathbf{\hat{x}}$ the prediction, $\mathbf{x}$ the true observations, the three scores are defined as
\begin{equation}
\begin{split}
    {RMSE}(\mathbf{x},\mathbf{\hat{x}}) = \sqrt{ \frac{1}{|\mathrm{J}|} \sum_{j\in \mathrm{J}} (x_j - \hat{x}_j)^2  }, \\    {MAE}(\mathbf{x},\mathbf{\hat{x}})=\frac{1}{|\mathrm{J}|} \sum_{j\in \mathrm{J}} \left|x_j - \hat{x}_j\right|, \\
    R^2(\mathbf{x},\mathbf{\hat{x}}) = 1 - \frac{ \sum_{j\in \mathrm{J}} (x_j - \hat{x}_j)^2} {\sum_{j\in \mathrm{J}} (x_j - \text{mean}(\mathbf{x}))^2}.
    \end{split}
\end{equation}

\subsection{Datasets}

We use two publicly available delay data to experiment with network-wide airport delays of different characteristics. The first U.S. delay dataset is collected from the U.S. Bureau of Transportation Statistics (BTS) database (\url{https://www.transtats.bts.gov/DL SelectFields.asp?gnoyr VQ=}). Seven-year flight data from January 1st, 2015, to December 31st, 2021, are collected. The initially collected dataset includes 360 airports. We select 70 airports with heavier traffic volumes for our experiments. The U.S. weather dataset of the same airports during the same time period is obtained from~\cite{moosavi2019short}. Eight weather categories, which include normal weather, severe cold, fog, hail, rain, snow, storm, and other precipitation, are considered. The second China delay dataset is collected from Xiecheng (\url{https://pan.baidu.com/s/1dEPyMGh#list/path=\%2F}). Two-year flight data from April 30th, 2015, to May 1st, 2017, are collected. Seven weather categories, which include normal weather, rain, cloud, thunderstorm, fog, storm and snow, are obtained from the associated special event data. Only flight records between 6 am to 12 pm are considered for those two datasets because very few flights are observed outside this period.

We aggregate the flights’ arrival and departure delays into 30 minutes windows according to their origin and destination in both datasets. Consequently, the exact value of flights with delays higher than 30 minutes can not be obtained in real-time. We use the following equation to compute the average arrival and departure delay ($a_{v, k}$, $d_{v, k}$) of airport $v$ at time point $k$:
\begin{equation}
\begin{split}
    a_{v, k} = \frac{\sum_{i \in \mathrm{V}} \sum_{j \in \mathrm{K}} \text{min}(d^a_{i, j}, 30)}{|\mathrm{V} | | \mathrm{K}|}, \\
    d_{v, k} = \frac{\sum_{i' \in \mathrm{V'}} \sum_{j' \in \mathrm{K'}} \text{min}(d^d_{i', j'}, 30)}{|\mathrm{V'} | | \mathrm{K'}|},
    \end{split}
    \label{eq:delay_form}
\end{equation}
where $\mathrm{V}$ and $\mathrm{V'}$ are the set of flights whose destination and origin are at airport $v$, respectively, $\mathrm{K}$ and $\mathrm{K'}$ are the set of flights whose schedule arrival and departure time are during $[k, k+30)$, $d^a_{i, j}$ and $d^d_{i', j'}$ is the corresponding flights’ arrival and departure delays. The form in Equation~\eqref{eq:delay_form} directly illustrates the delay level, with a maximum value of 30 min. After aggregation, the U.S. delay dataset contains 16\% unobserved data (missing data or data point without a flight), the China delay dataset contains 42\% unobserved data.   

To construct the multi-relational graph of airports, we use three types of adjacency matrices. Like typical spatiotemporal prediction models \cite{li2018diffusion, yu2018spatio}, we compute the pairwise distances between airports and build the distance adjacency matrix $A_d$ using thresholded Gaussian kernel $A_d(i, j) = exp\left(-\frac{dist(v_i, v_j)^2}{\sigma^2}\right)$, and we set $A_d(i, j) = 0$ if $A_d(i, j) \leq 0.1$. In addition, we also consider the origin-destination (O-D) and destination-origin (D-O) relations. The O-D adjacency matrix is computed by:
\begin{equation}
    A_{O\to D}(i, j) = \begin{cases}
    0, \quad & \text{if} \quad F_{i \to j} < 0.15\hat{F}_{O \to D}, \\
    \frac{F_{i \to j}}{1.5\hat{F}_{O \to D}}\quad & \text{otherwise},
    \end{cases}
    \label{eq:od_adj}
\end{equation}
where $F_{i \to j}$ is the total air flow from airport $i$ to $j$ of the training dataset, $\hat{F}_{O \to D}$ is the maximum value of OD flow pair in the training dataset. The destination-origin adjacency matrix $ A_{D\to O}$ can be directly computed by $A_{O\to D}^T$.

\subsection{Experimental Setups}

We conduct multi-step ahead delay prediction for both two datasets. For the U.S. delay dataset, we use 12 previous time points (6 hours) to predict delays of 12 future time points (6 hours). For the China dataset, we use 36 previous time points (18 hours) to predict delays of 12 future time points (6 hours) because it contains more missing data. Z-score normalization is applied to inputs. The datasets are split in chronological order with 70\% for training, 10\% for validation and 20\% for testing.

We compare STPN with widely used spatiotemporal prediction models, including (1) HA: Historical Average, which models the airport delay as a seasonal process, and uses the weighted average of previous seasons as the prediction. The period used is one week, and the prediction is based on aggregated data from previous weeks. (2) VAR: Vector Auto-Regression. The lags are set to 12 for the U.S. delay dataset and 36 for the China delay dataset. (3) Long Short-term Memory (LSTM): The Encoder-decoder framework using LSTM with a peephole. Both the encoder and the decoder contain two recurrent layers. In each recurrent layer, there are 256 units. The model is trained with batch size 64 and loss function MAE, the learning rate is 0.01, and the early stop is performed by monitoring the validation error. The LSTM model does not utilize spatial correlation. We train one LSTM model to predict departure and arrival delays for all airports. (4) Spatiotemporal Graph Convolutional Network (STGCN) \cite{yu2018spatio}: The graph convolution layer of STGCN only contains one graph. We use the O-D adjacency matrix given in Equation~\eqref{eq:od_adj} to perform graph convolution. The channels of three layers in ST-Conv block of STGCN are 64, 16, and 64, respectively. The graph convolution kernel size and temporal convolution kernel size are set to 3 in the model. We train STGCN by minimizing the root mean square error using Adam for 50 epochs with a batch size of 32. The learning rate is 0.001. (5) Graph Wavenet (Gwave) \cite{wu2019graph}. We use eight layers of Graph WaveNet. We use Equation~\eqref{eq:random_walk} as our graph convolution layer with a diffusion step 2. We train Gwave using Adam optimizer with an initial learning rate of 0.001. The STGCN and Gwave models are originally used for single variable forecasting. We replace their output layer with a fully-connected layer that outputs two variables. 

We implement the STPN model based on the PyTorch\footnote{https://pytorch.org/} framework. We choose Root Mean Square error as the loss function. The hyperparameters and the best models are determined by the performance of the validation sets. We use four layers of space-time-separable multi-graph convolution in Equation~\eqref{eq:tensor_multiple_graph}, and the last layer is the output layer. The feature dimension of the hidden layer is [128, 64, 32]. Each hidden layer is followed by a SE block with the reduction rate $r = 16$. The number of the head for the attention model is 4. The order of graph convolution is 2. We use a 4-dimensional embedding to encode weather data. We train STPN using Adam optimizer with a learning rate of 0.001.

\subsection{Main Results and Analysis}

\begin{table*}[htb]
  \caption{Results on the U.S. delay dataset}
  \footnotesize
  \label{sample-table_US}
  \centering
  \begin{tabular}{cllll|lll|lll}
    \toprule
  &   & \multicolumn{3}{c}{1.5 hour} &  \multicolumn{3}{c}{3 hour}  & \multicolumn{3}{c}{6 hour}               \\
 &{Method} & $MAE$     & $RMSE$     & $R^2$ & $MAE$     & $RMSE$     & $R^2$ & $MAE$     & $RMSE$     & $R^2$ \\
    \midrule
{\multirow{6}{*}{Arrival  Delay}}&HA & 9.089 & 11.847 & 0.040 & 9.089 & 11.847 & 0.040 & 9.089 & 11.847 & 0.040 \\
&VAR &7.795 &10.468 &0.251 &8.123 &10.824 &0.199   &8.479 &11.237 &0.136 \\
&LSTM &7.979 &10.751 &0.209 &8.354 &11.153 &0.149  &8.681 &11.500 &0.073 \\
&STGCN &7.784 &10.509 &0.248 &7.928 &10.630 &0.233  &8.157 &10.878 &0.199 \\
&Gwave &7.771 &10.458 &0.252 &7.860 &10.563 &0.237   &8.085 &10.819 &0.200 \\
&STPN &\textbf{7.202} &\textbf{9.806} &\textbf{0.342}   &\textbf{7.502} &\textbf{10.159} &\textbf{0.294}     &\textbf{7.780} &\textbf{10.552} &\textbf{0.239} \\
    \midrule 
{\multirow{6}{*}{{Departure Delay}}}&HA &6.519  &8.632  &0.069 &6.519  &8.632  &0.069 &6.519  &8.632  &0.069   \\
&VAR &5.560 &7.655 &0.268 &5.816 &7.924 &0.216  &6.165 &8.303 &0.139 \\
&LSTM &5.509 &7.846 &0.230 &5.798 &8.195 &0.161  &6.115 &8.616 &0.096 \\
&STGCN &5.391 & 7.605 &0.277 &5.540 &7.692 &0.261  &5.856 &7.998 &0.200 \\
&Gwave &5.437 &7.624 &0.273 &5.507 & 7.701 &0.259   &5.705 &7.862 &0.228 \\
&STPN &\textbf{5.217} &\textbf{7.430} &\textbf{0.310}   &\textbf{5.320} &\textbf{7.528} &\textbf{0.292}     &\textbf{5.512} &\textbf{7.713} &\textbf{0.257} \\
    \bottomrule
  \end{tabular}
\end{table*}

\begin{table*}[htb]
  \caption{Results on the China delay dataset}
  \footnotesize
  \label{sample-table_China}
  \centering
  \begin{tabular}{cllll|lll|lll}
    \toprule
  &   & \multicolumn{3}{c}{1.5 hour} &  \multicolumn{3}{c}{3 hour}  & \multicolumn{3}{c}{6 hour}               \\
 &{Method} & $MAE$     & $RMSE$     & $R^2$ & $MAE$     & $RMSE$     & $R^2$ & $MAE$     & $RMSE$     & $R^2$ \\
    \midrule
{\multirow{6}{*}{Arrival  Delay}}&HA &10.720  & 13.123  &0.095  &10.720  & 13.123  &0.095 &10.720  & 13.123  &0.095  \\
&VAR &8.970 &11.372 &0.320 &9.339 &11.750 &0.274  &9.816 &12.244 &0.212 \\
&LSTM &9.047 &11.574 &0.295 &9.456 &12.023 &0.240   &10.218 &12.783 &0.162 \\
&STGCN &8.474 &\textbf{10.838} &\textbf{0.382} &9.065 &11.590 &0.293  &9.685 &12.083 &0.232 \\
&Gwave &9.279 &11.765 &0.272 &9.476 &11.999 &0.243  &9.895 &12.410 &0.190 \\
&STPN &\textbf{8.454} & 10.861 & 0.380  &\textbf{8.708} &\textbf{11.436} &\textbf{0.342} &\textbf{9.618} &\textbf{12.050} &\textbf{0.236} \\
    \midrule
{\multirow{6}{*}{{Departure Delay}}}&HA &10.441  & 12.929 &0.144 &10.441  & 12.929 &0.144 &10.441  & 12.929 &0.144   \\
&VAR &8.927 &\textbf{11.376} &\textbf{0.336} &9.250 &11.709 &0.297 &9.675 &12.164 &0.241\\
&LSTM &9.499 &12.014 &0.260 &9.850 &12.385 &0.213   &9.969 &12.569 &0.169 \\
&STGCN &9.017 &11.495  &0.322 &9.203 &11.716 &0.296  &\textbf{9.513} &12.068 &0.253 \\
&Gwave &9.246 &11.855 &0.279 &9.426 &12.060  &0.254  &9.897 &12.514 &0.197 \\
&STPN &\textbf{8.789} &11.381 &0.336 &\textbf{9.120} &\textbf{11.612} &\textbf{0.308} &9.775 &\textbf{12.007} &\textbf{0.260} \\
    \bottomrule
  \end{tabular}
\end{table*}

Table~\ref{sample-table_US} and \ref{sample-table_China} summarize the results for U.S. and China delay datasets. A few observations follow. (1) Graph neural networks, including STGCN, Gwave, and STPN, generally outperform other baselines on the U.S. delay dataset. This result corroborates the premise of this work: graph structure can be used to capture the spatiotemporal propagation patterns of airport delays. (2) Among the graph-based methods, STPN performs better than STGCN and Gwave. Unlike those two baselines, STPN utilizes the self-attention mechanism instead of temporal convolutional networks to capture temporal dependencies and employs the SE block to model feature relationships. Those characteristics make STPN better at capturing the spatiotemporal dependencies and arrival-departure interactions within delay datasets. (3) The China delay dataset is significantly more challenging than the U.S. delay dataset for the deep learning method. Gwave does not give satisfactory on this dataset, and the reason might be that its structure does not process missing data very well. STPN still outperforms other baselines on long-range forecasting, illustrating the advantage of the attention mechanism on long-range temporal dependencies modeling. Surprisingly VAR performs reasonably well despite simplicity on the China delay dataset. It indicates that a relatively simple model might be less vulnerable to missing data. (4) LSTM does not give satisfactory results on both the U.S. and China delay datasets. Its performance is even worse than VAR's, indicating that recurrent neural networks are not suitable for handling long-range temporal dependencies within airport flight delays.

\begin{figure*}[!ht]
\centering
\subfigure[Spatial visualization of 3-step ahead arrival delay prediction on U.S. dataset]{
\includegraphics[width = 0.5\textwidth]{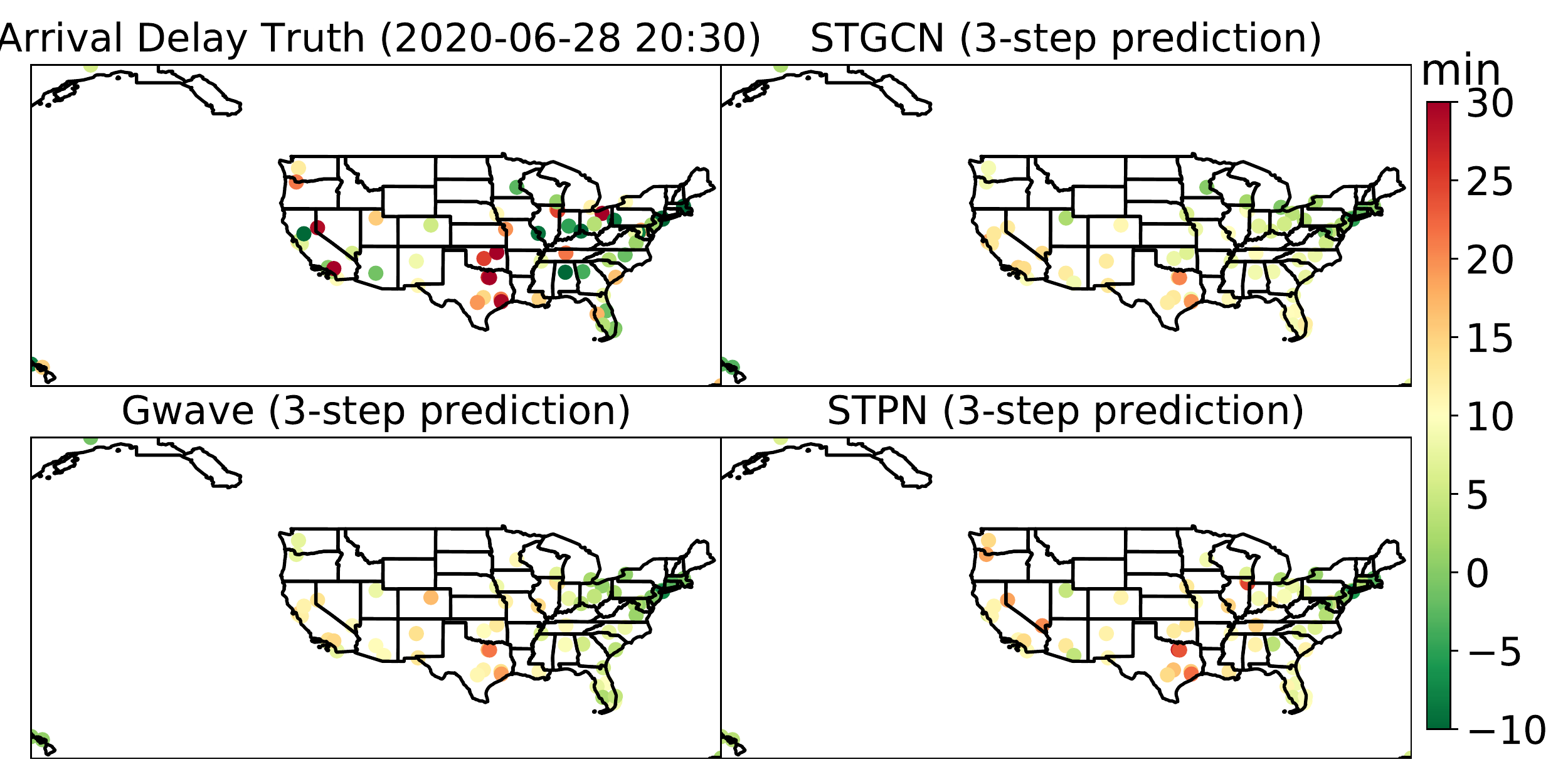}
\label{fig:us_space}}
\subfigure[Spatial visualization of 12-step ahead arrival delay prediction on China dataset]{ 
\includegraphics[width = 0.36\textwidth]{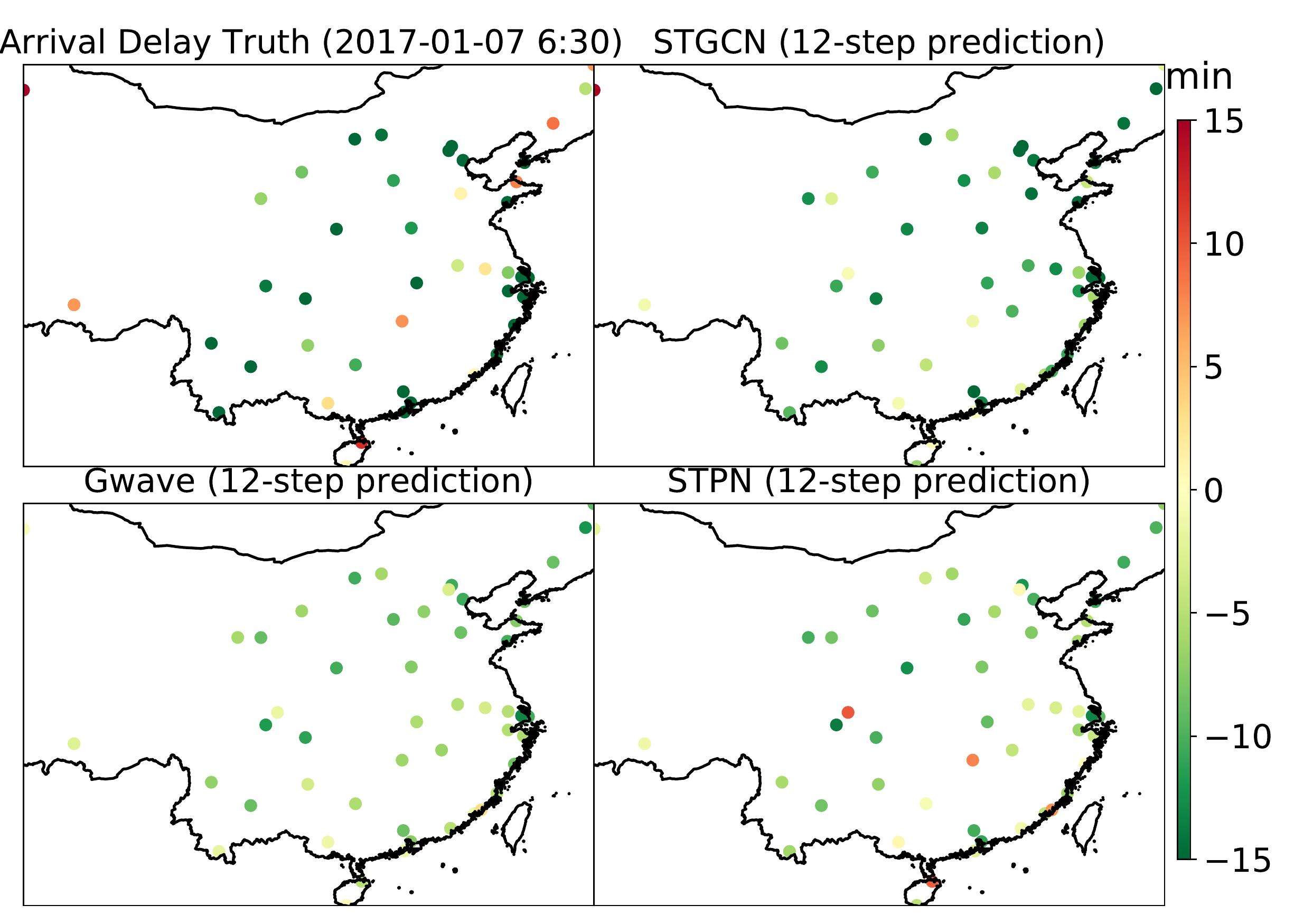}
\label{fig:uch_space}}
\subfigure[Temporal visualization of arrival delay prediction on ORD airport from U.S dataset]{
\includegraphics[width = 0.43\textwidth]{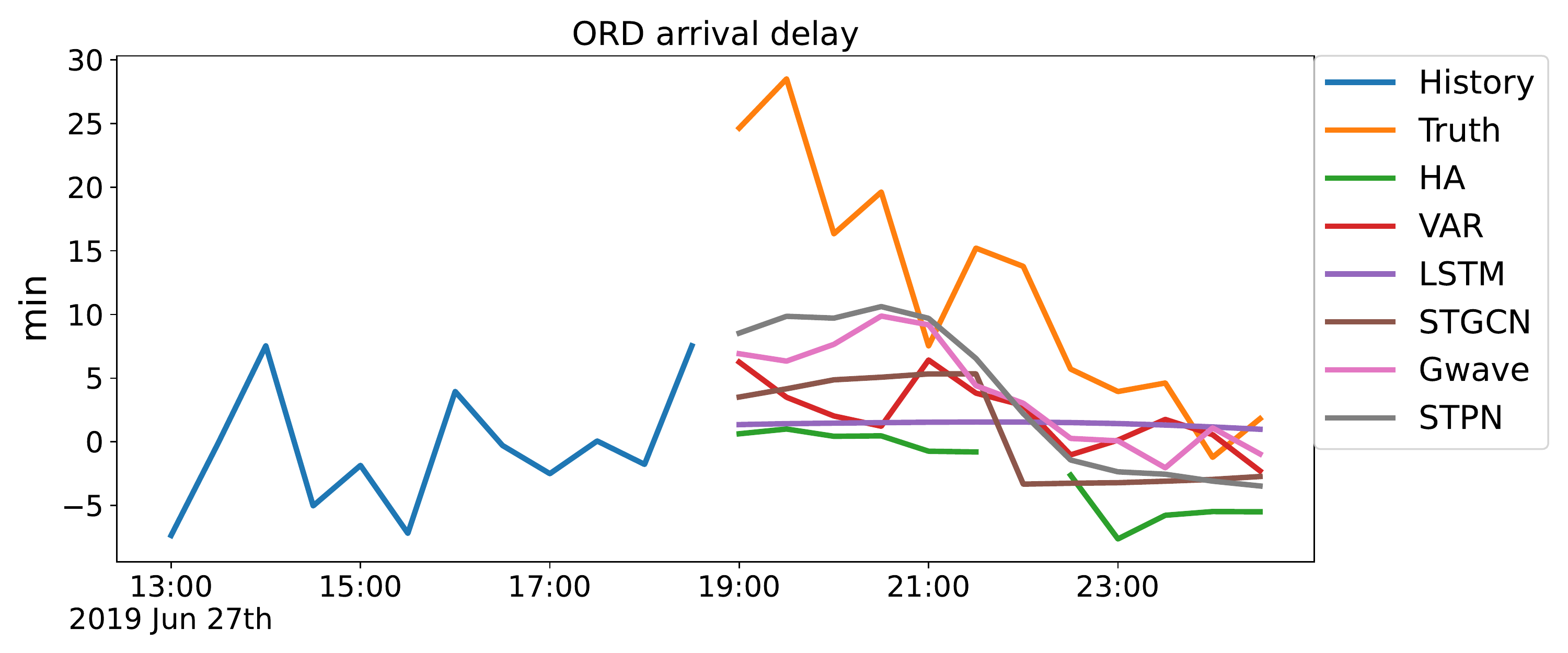}
\label{fig:us_time}}
\subfigure[Temporal visualization of departure delay prediction on HGH airport from China dataset]{ 
\includegraphics[width = 0.43\textwidth]{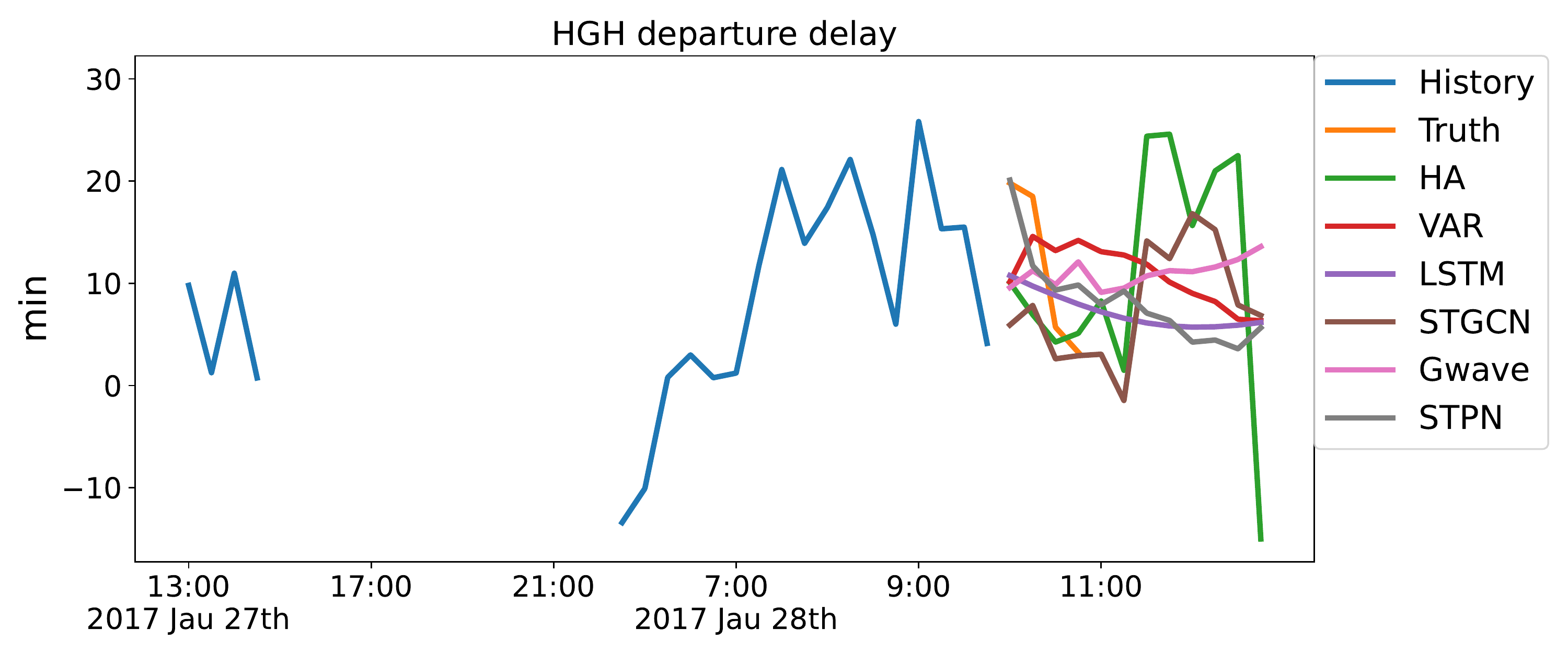}
\label{fig:uch_time}}
    \caption{Qualitative visualization of prediction results.}
\end{figure*}

Figure~\ref{fig:us_space} gives the spatial visualization of 3-step ahead arrival delay prediction on U.S. dataset. It is clear that STPN model produces the closest estimation toward true values compared with STGCN and Gwave. Furthermore, STPN better approximates the high arrival delay at Northwestern airports with the learned long-range temporal dependencies. Figure~\ref{fig:uch_space} shows the spatial visualization of 12-step ahead (6 hours) arrival delay prediction on China dataset. STPN accurately predicts the high arrival delay at HHA airport (Central red value) 6 hours ahead, while other models fail. To qualitatively illustrate the performance of STPN on multi-step ahead prediction, we also visualize the temporal results in Figure~\ref{fig:us_time} and ~\ref{fig:uch_time}. Although all models fail to estimate the sudden rise of arrival delay at ORD airport in Figure~\ref{fig:us_time}, STPN still gives the best results compared with other baselines. The missing data ratio of HGH departure delay in Figure~\ref{fig:uch_time} is relatively high. However, STPN can still give the most accurate results compared with other baselines. Using those examples, we demonstrate that STPN can be potentially used in practice to perform air traffic flow control \cite{menon2004new}, given the ability to predict arrival/departure delays several hours ahead. 

\subsection{Visualization of Temporal Attention Matrix}

\begin{figure*}[!htb]
\centering
\subfigure[]{
\includegraphics[width = 0.23\textwidth]{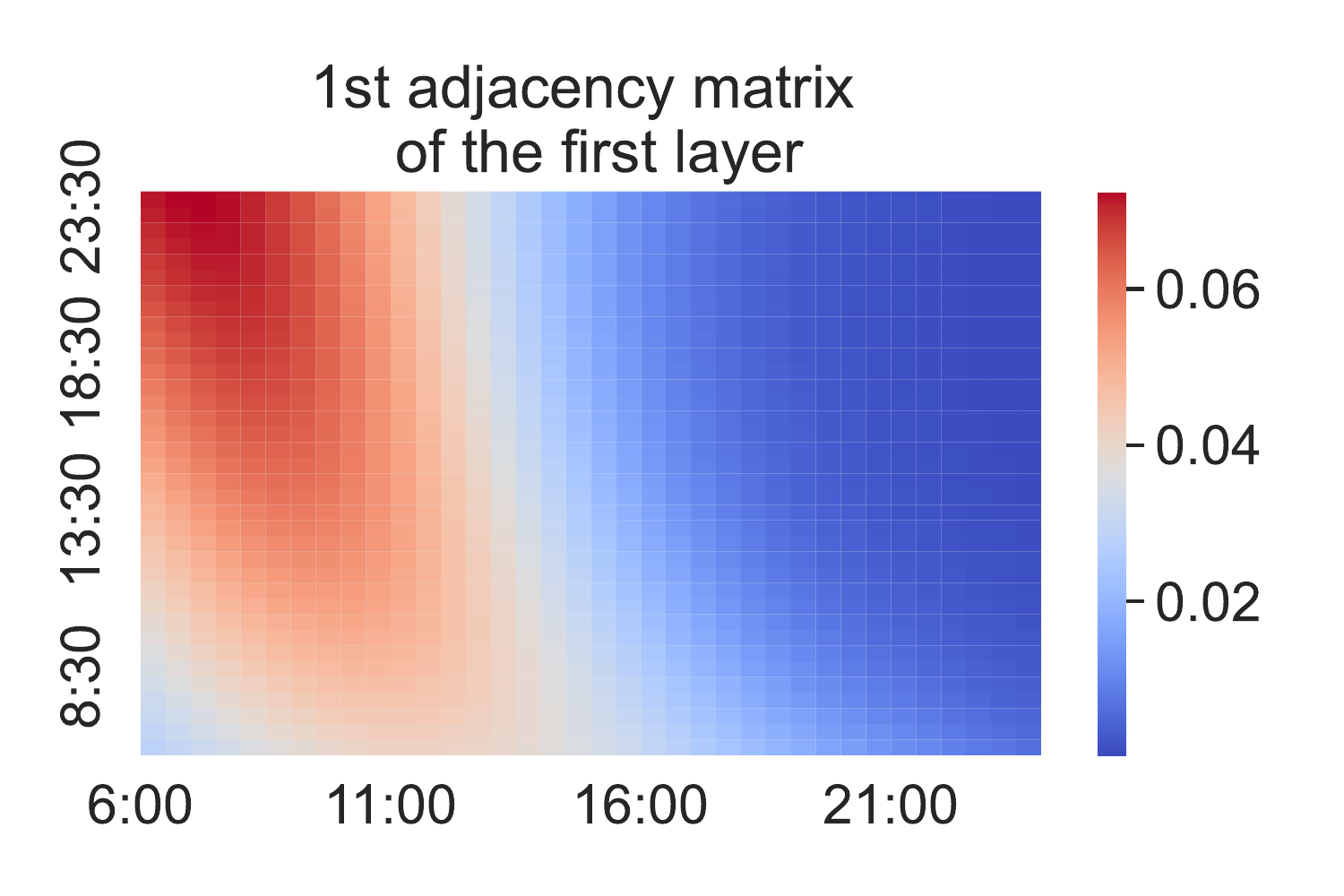}}
\subfigure[]{
\includegraphics[width = 0.23\textwidth]{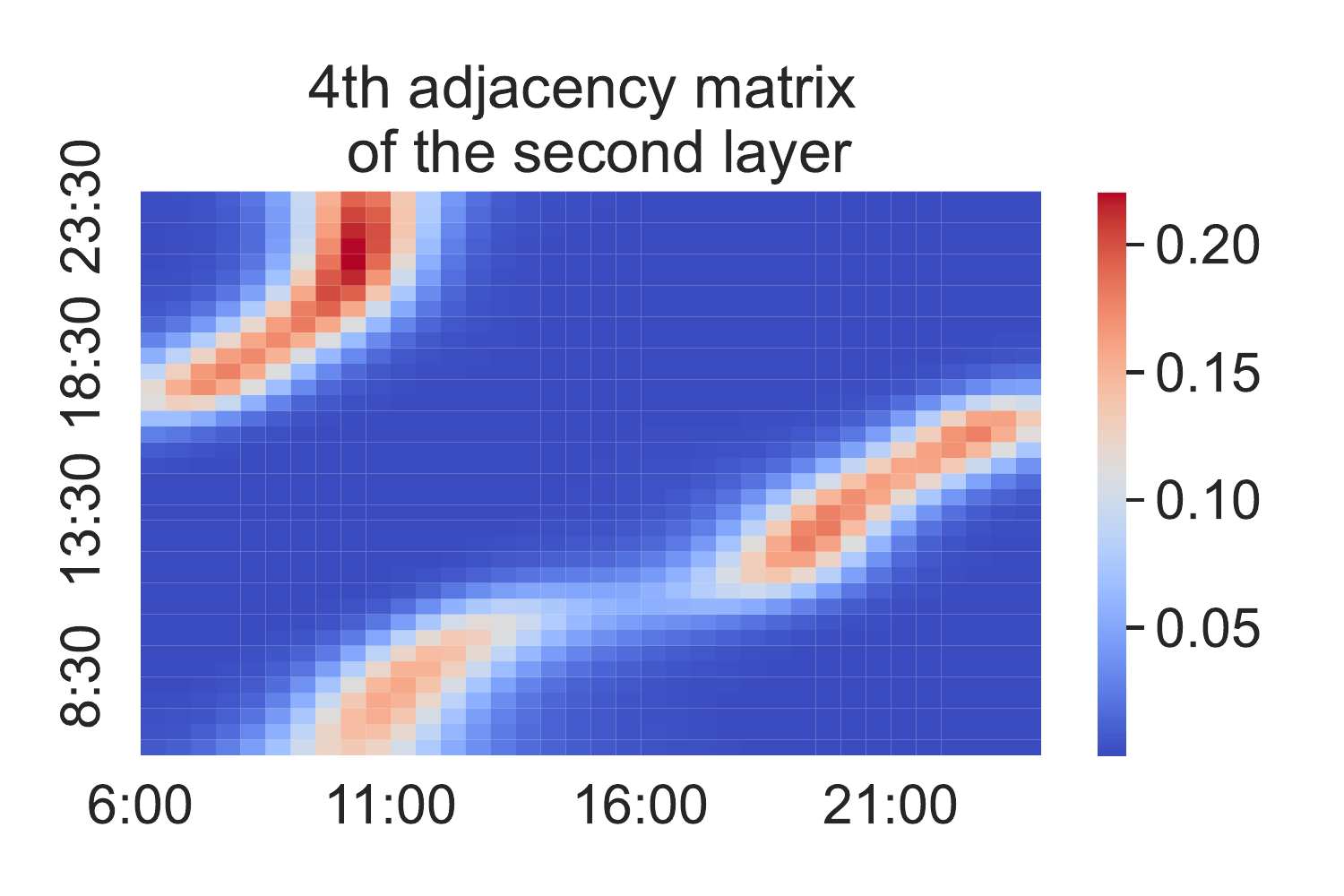}}
\subfigure[]{
\includegraphics[width = 0.23\textwidth]{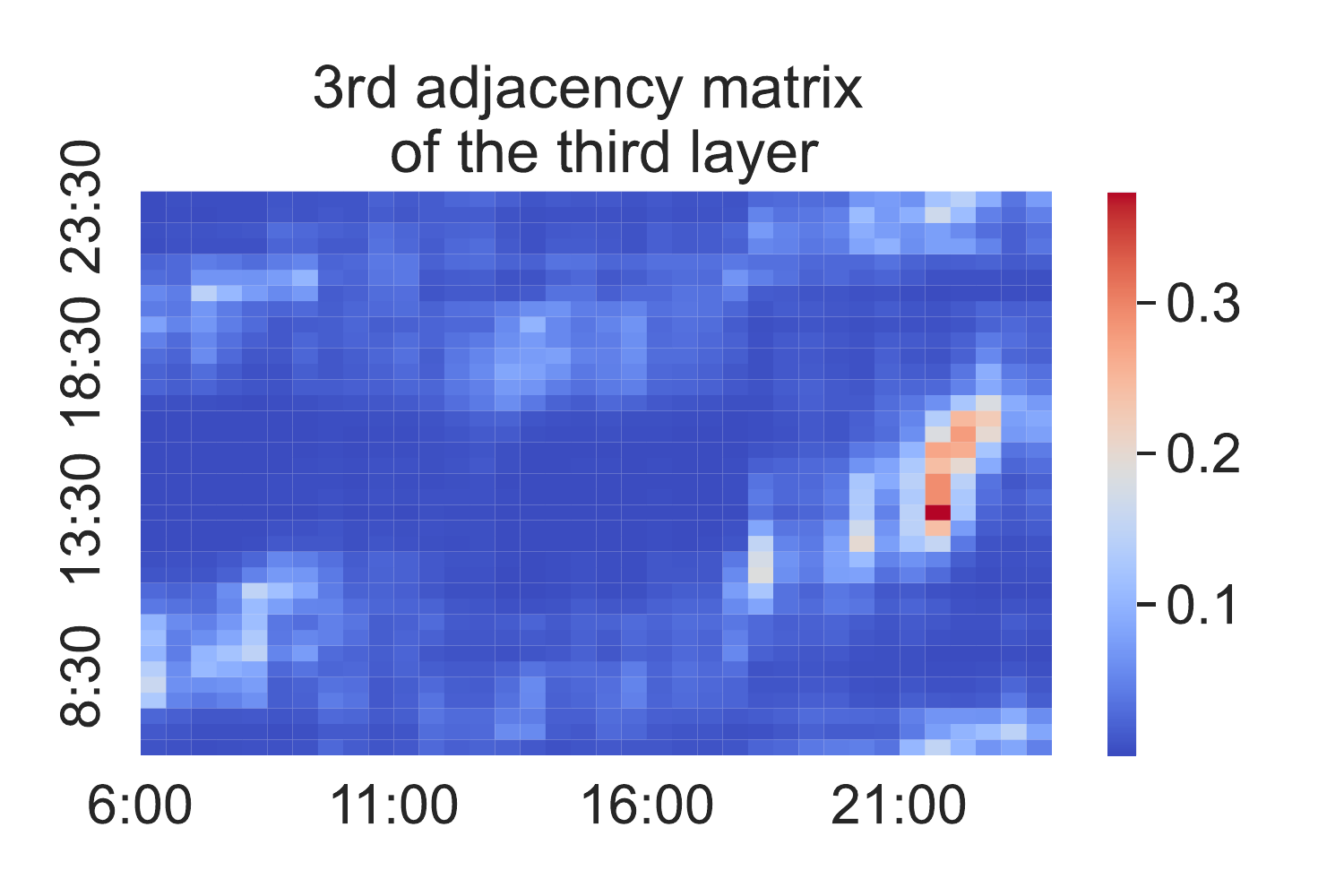}}
\subfigure[]{
\includegraphics[width = 0.23\textwidth]{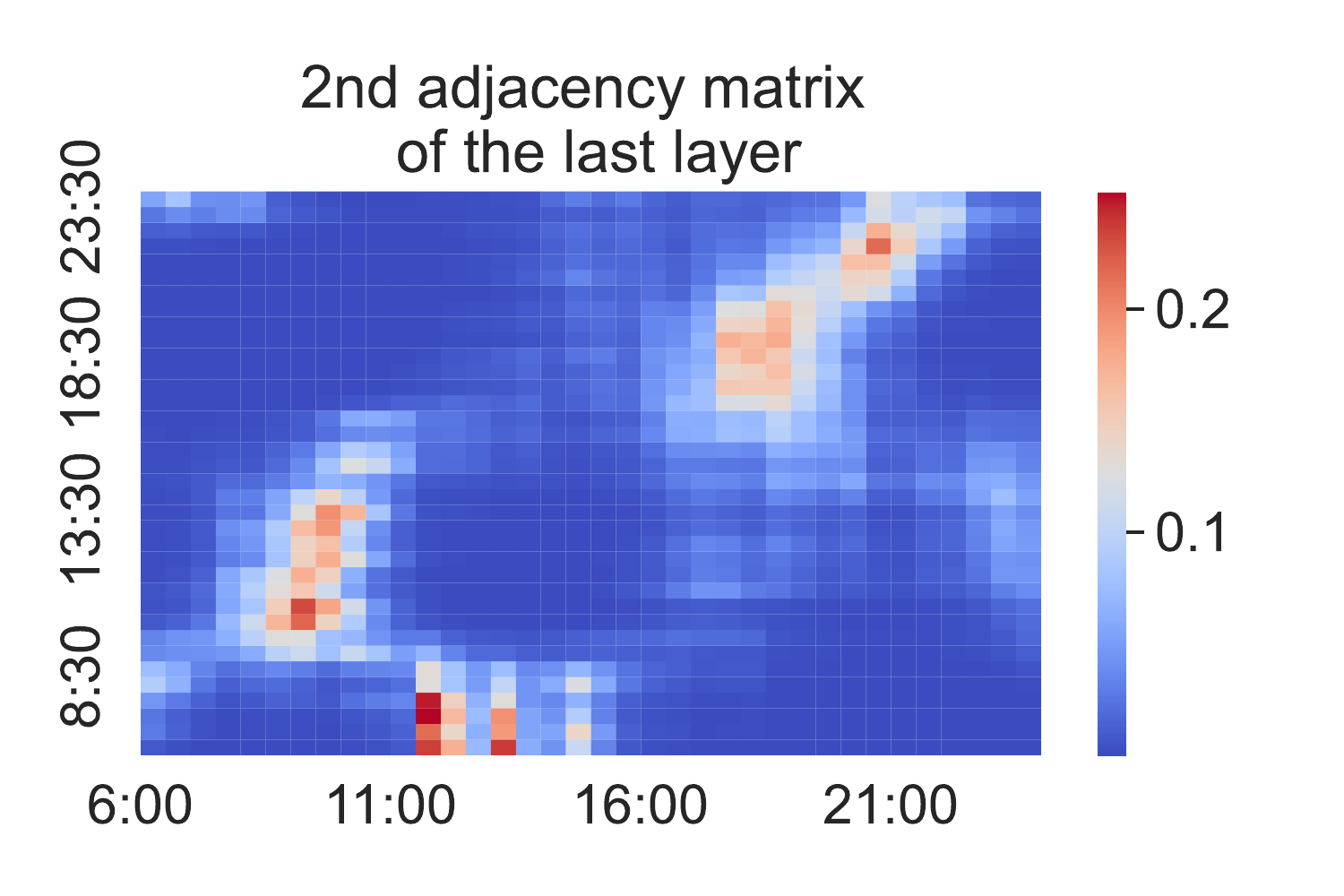}}
\subfigure[]{
\includegraphics[width = 0.23\textwidth]{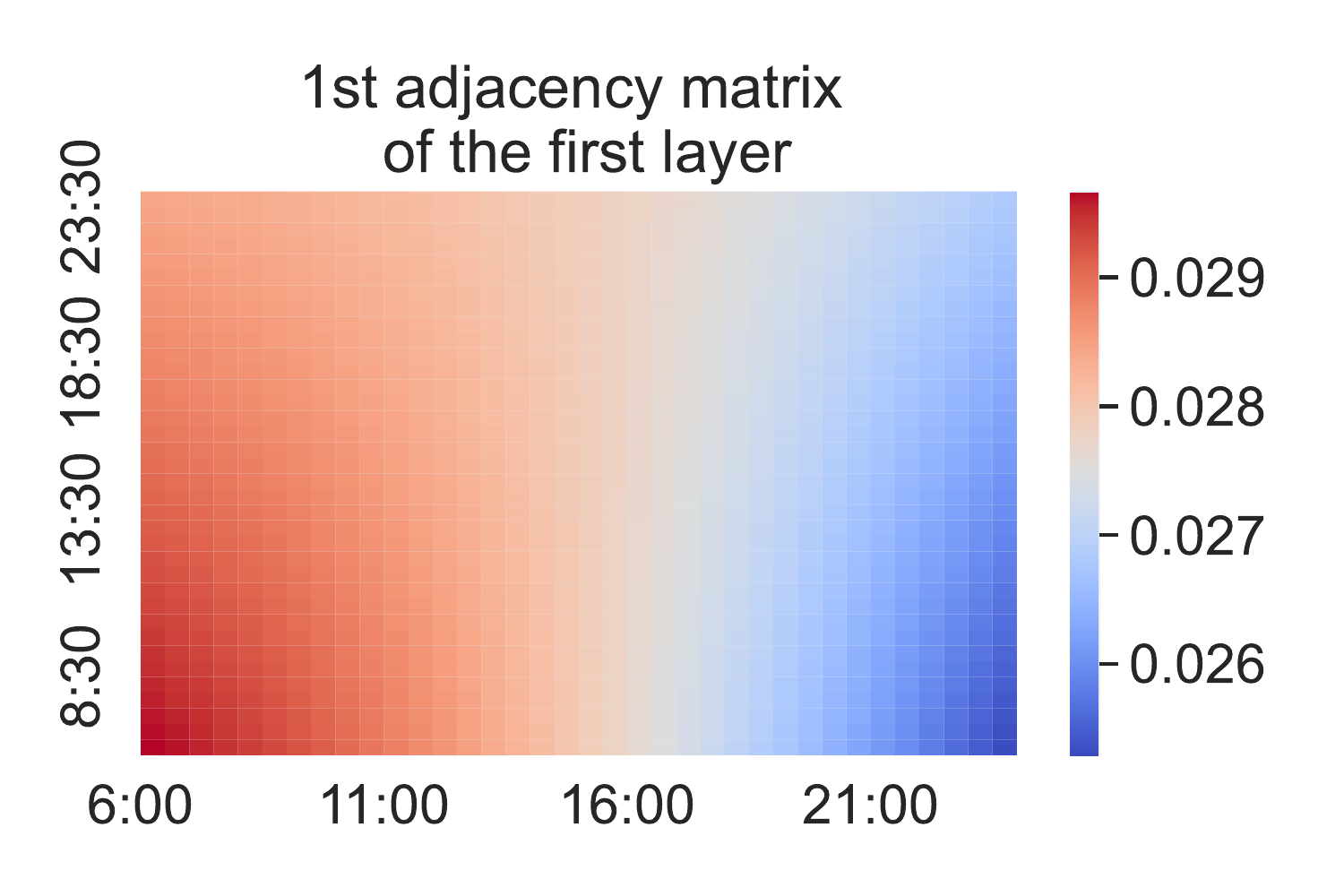}}
\subfigure[]{
\includegraphics[width = 0.23\textwidth]{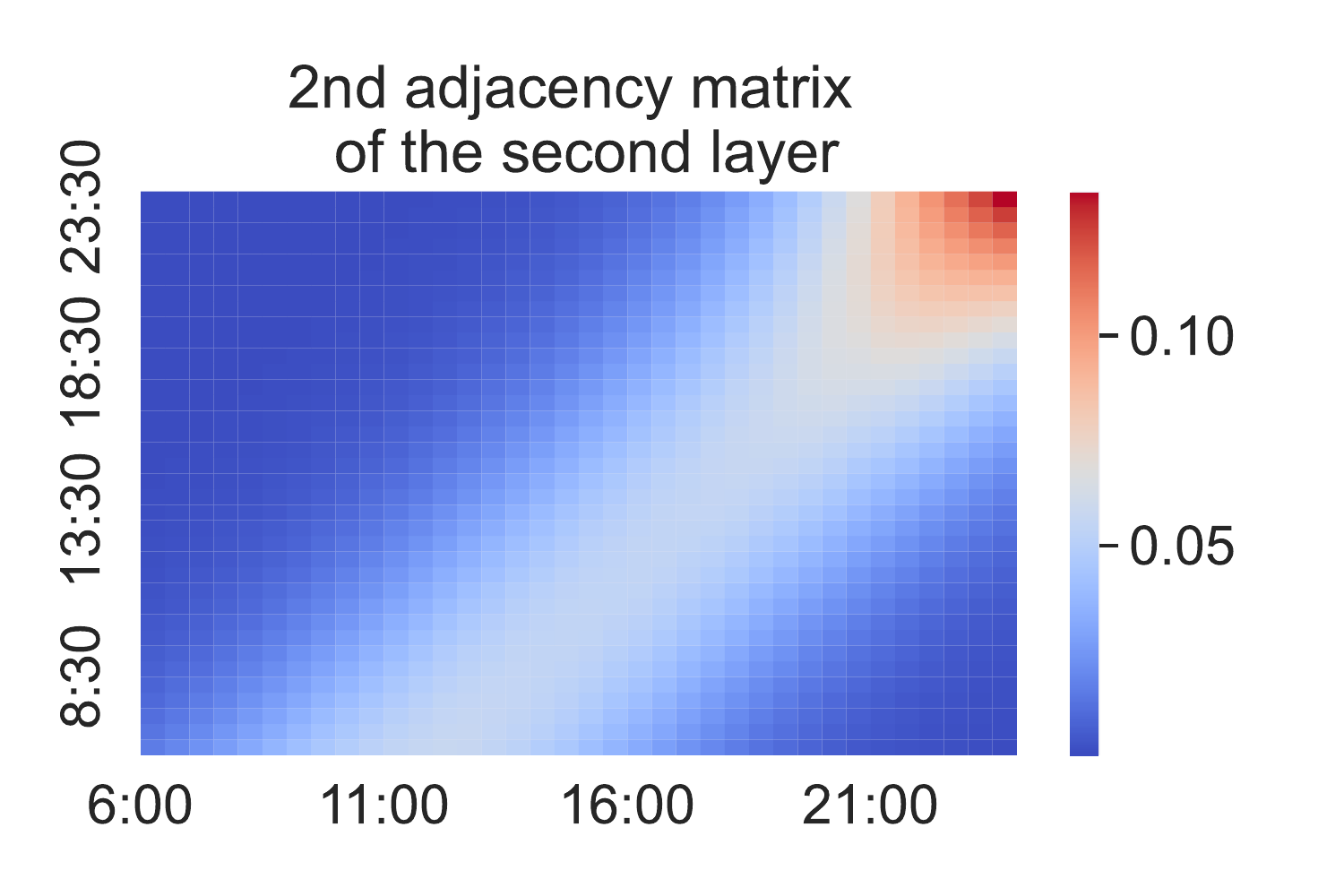}}
\subfigure[]{
\includegraphics[width = 0.23\textwidth]{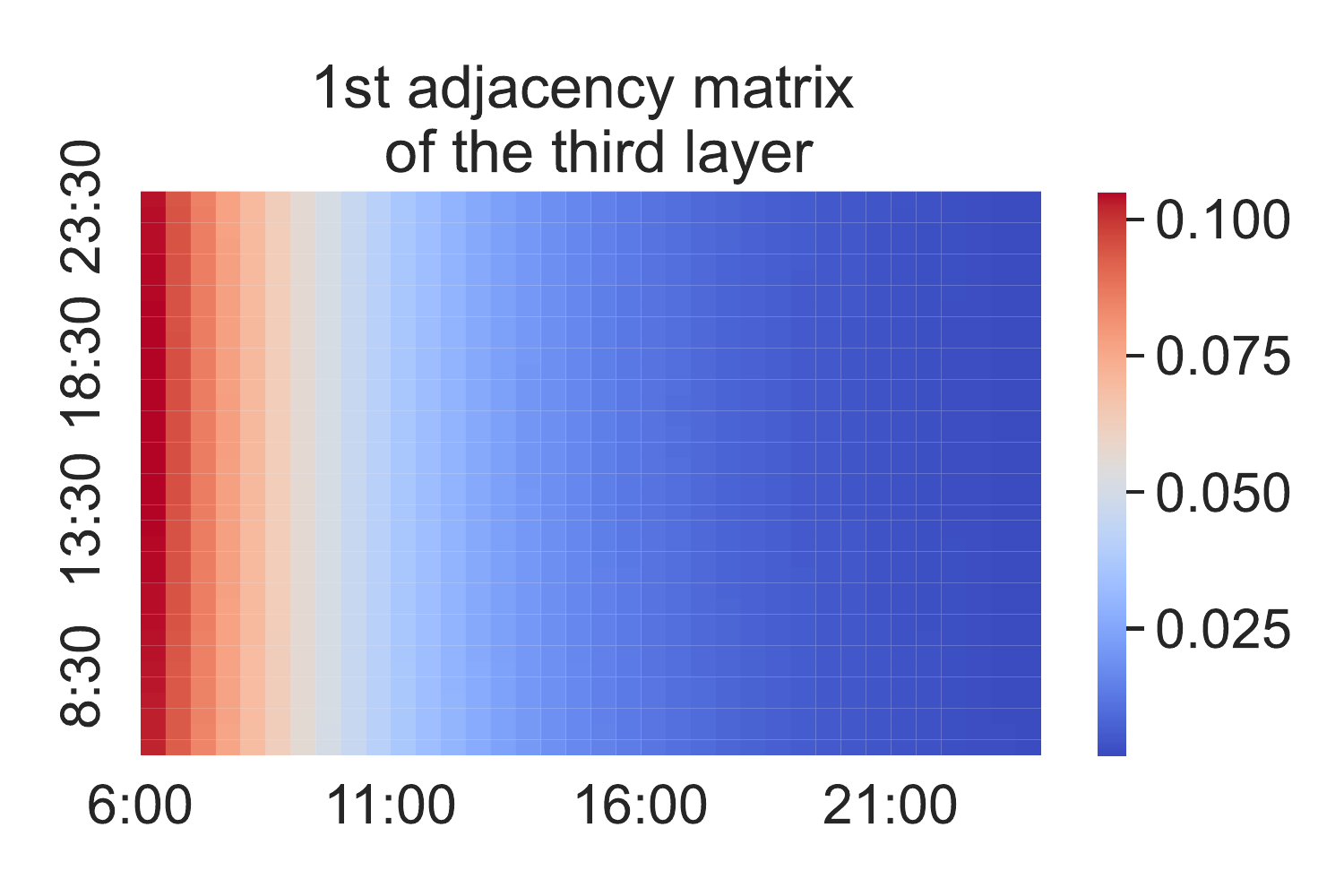}}
\subfigure[]{
\includegraphics[width = 0.23\textwidth]{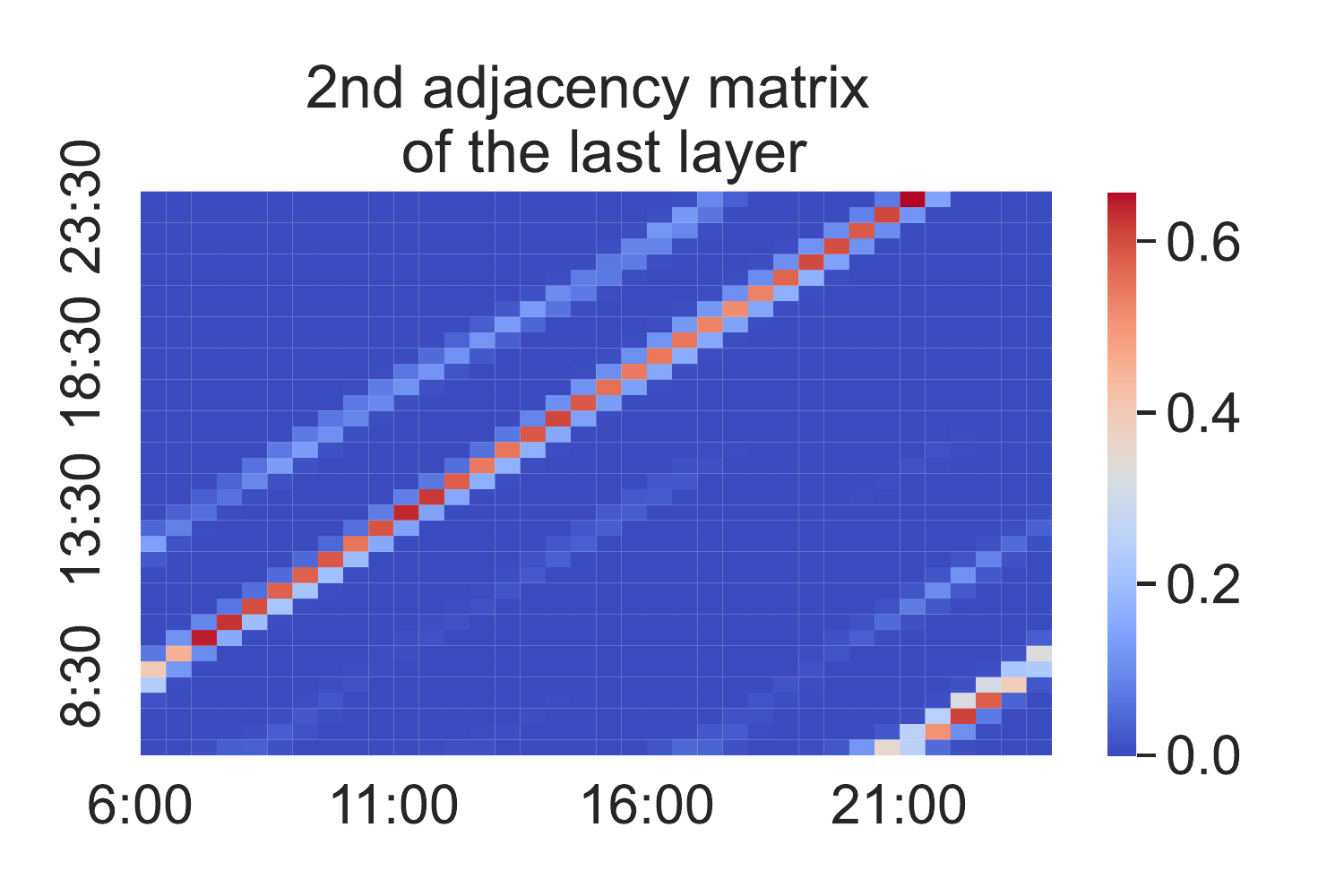}}
    \caption{Raw Attention maps of STPN at different layers.}
    \label{fig:attention}
\end{figure*}

The main STPN module for capturing temporal dependencies are self-attention layers, which assign a pairwise attention score between every two points. Although it is wrong to equate attention scores with explanation \cite{jain2019attention}, it can offer plausible and meaningful interpretations \cite{li2019enhancing,chefer2021transformer}. To understand how STPN learns temporal dependencies from delay datasets, we calculate each layer's attention scores between every time point of the day. We visualize parts of raw attention scores $A^{(l)}_T = {\operatorname{softmax}}\left ( {\frac {Q^{(l)}(K^{(l)})^{\intercal  }}{\sqrt {c_{k}}}} \right )^\intercal$ in Figure~\ref{fig:attention}. Crucially, every attention matrix in different layers learns distinct temporal attention patterns, reflecting that delays in an airport network contain very complex temporal dependencies. Another observation is that we learn more sparse attention scores as we go deeper into the model. For instance, the second adjacency matrix of the last layer of STPN trained on the China dataset only gives high scores between time points with 2.5-hour delays. This pattern resonates with the fact that the travel time of most Chinese flights ranges from 1.5 hours to 3 hours. It seems that every temporal attention in deeper layers is only responsible for one type of temporal dependency. We also find that most attention maps in the deeper layers only give high attention scores between temporally closer time points. Only the attention maps in the first layer give uniform attention scores between every time point. It means that the self-attention model of the first STPN layer learns some global temporal dependencies, while deeper layers learn local dependencies. It shows that STPN uses both short-range and long-range temporal dependencies for delay prediction.

\subsection{Ablation Analysis}

\begin{table*}[htb]
  \caption{Ablation analysis on the U.S. delay dataset (The best indexes except the ones of STPN are marked in bold, and the worst ones are marked in red.)}
  \footnotesize
  \label{sample-table_ablation}
  \centering
  \begin{tabular}{l|lll|lll|lll}
    \toprule
    & \multicolumn{3}{c}{overall metrics} &  \multicolumn{3}{c}{Arrival delay}  & \multicolumn{3}{c}{Departure Delay}               \\
 {Method} & $MAE$     & $RMSE$     & $R^2$ & $MAE$     & $RMSE$     & $R^2$ & $MAE$     & $RMSE$     & $R^2$ \\
    \midrule
STPN &6.428 &8.967 &0.356  &7.506 &10.165 &0.293 &5.333 &7.542 &0.289 \\
STPN-WS &6.456 &9.017 &0.349  &\textbf{7.506} &10.193 &0.289 &5.350 &7.581 &0.282 \\
STPN-WC &6.457 &\textbf{8.996} &\textbf{0.352}  &7.523 &\textbf{10.185} &\textbf{0.290} &\textbf{5.333} &\textbf{7.543} &\textbf{0.289} \\
STPN-WOD &6.448 &9.004 &0.350  &7.508 &10.197 &0.289 &5.334 &7.544 &0.289 \\
STPN-WDis &\textbf{6.442} &9.007 &0.350  &7.507 &10.188 &0.290 &5.338 &7.564 &0.285 \\
STPN-WG &6.487 &9.065 &0.342  &7.559 &10.253 &0.281 &{\color{red}5.355} &{\color{red}7.613} &{\color{red}0.276} \\
STPN-S &{\color{red}6.521} &{\color{red}9.126} &{\color{red}0.333}  &{\color{red}7.650} &{\color{red}10.392} &{\color{red}0.261} &5.340 &7.564 &0.285 \\
    \bottomrule
  \end{tabular}
\end{table*}

To understand the importance of the modules in STPN, we consider six ablations, and evaluate them on the U.S. delay dataset since it contains less missing data:
\begin{enumerate}
    \item \textbf{STPN-WS}: We remove the SE-block in each layer of STPN.
    \item \textbf{STPN-WC}: We remove the covariate (weather condition) inputs and their associated embeddings.
    \item \textbf{STPN-WOD}: We remove $A_{D\to O}$ and $A_{O\to D}$, and only use the distance adjacency matrix $A_{d}$ to perform graph convolution.
    \item \textbf{STPN-WDis}: We remove the distance adjacency matrix $A_{Dis}$.
    \item \textbf{STPN-WG}: We remove all the diffusion graph convolution blocks, which means that the spatial dependencies are not considered.
    \item \textbf{STPN-S}: Departure and arrival delays are trained separately, so the departure-arrival delay relationship is not considered.
\end{enumerate}

The results in Table~\ref{sample-table_ablation} show the performance of these ablations. The best indexes except the ones of STPN are marked in bold, and the worst ones are marked in red. The results show that our design yields performance improvements on both arrival and departure delays. Next, we describe the influence of each component individually.

\noindent \textbf{Improvements in SE-blocks}: The performance of STPN-WS is relatively worse than those of STPN-WC, STPN-WOD, and STPN-WDis, demonstrating the importance of SE-block in STPN. The SE-block helps capture the relationships between different features originating from arrival/departure delays and covariates. Those relationships are essential for delay prediction.  

\noindent \textbf{Improvements in covariate feature embedding}: The advantage of STPN over STPN-WC is marginal, indicating the contribution of covariate is limited in STPN. The reason might be that the effects of covariate have been reflected in historical delay patterns, and STPN can almost learn those effects by itself.

\noindent \textbf{Improvements in multi-graph convolution}: The performance of STPN-WOD and STPN-WDis are slightly worse than those of STPN, indicating that STPN can use one type of relation to infer sufficient spatial dependencies. However, STPN-WG performs worst in predicting departure delay, suggesting that spatial dependencies are more helpful in predicting departure delay. It is reasonable since the spatial departure delays are more likely local-correlated due to extreme weather conditions.

\noindent \textbf{Improvements in co-training}: Unsurprisingly, STPN-S gives the worst overall performance and arrival delay prediction performance. It is because arrival delay is highly related to departure delay. In many cases, departure delays are the cause of arrival delays. However, the impacts of arrival delays on departure delays are relatively small. Thus the departure delay prediction performance of STPN-S is even better than those of STPN-WG and STPN-WS.

\begin{figure}[!ht]
    \centering
    \includegraphics[width = 0.4\textwidth]{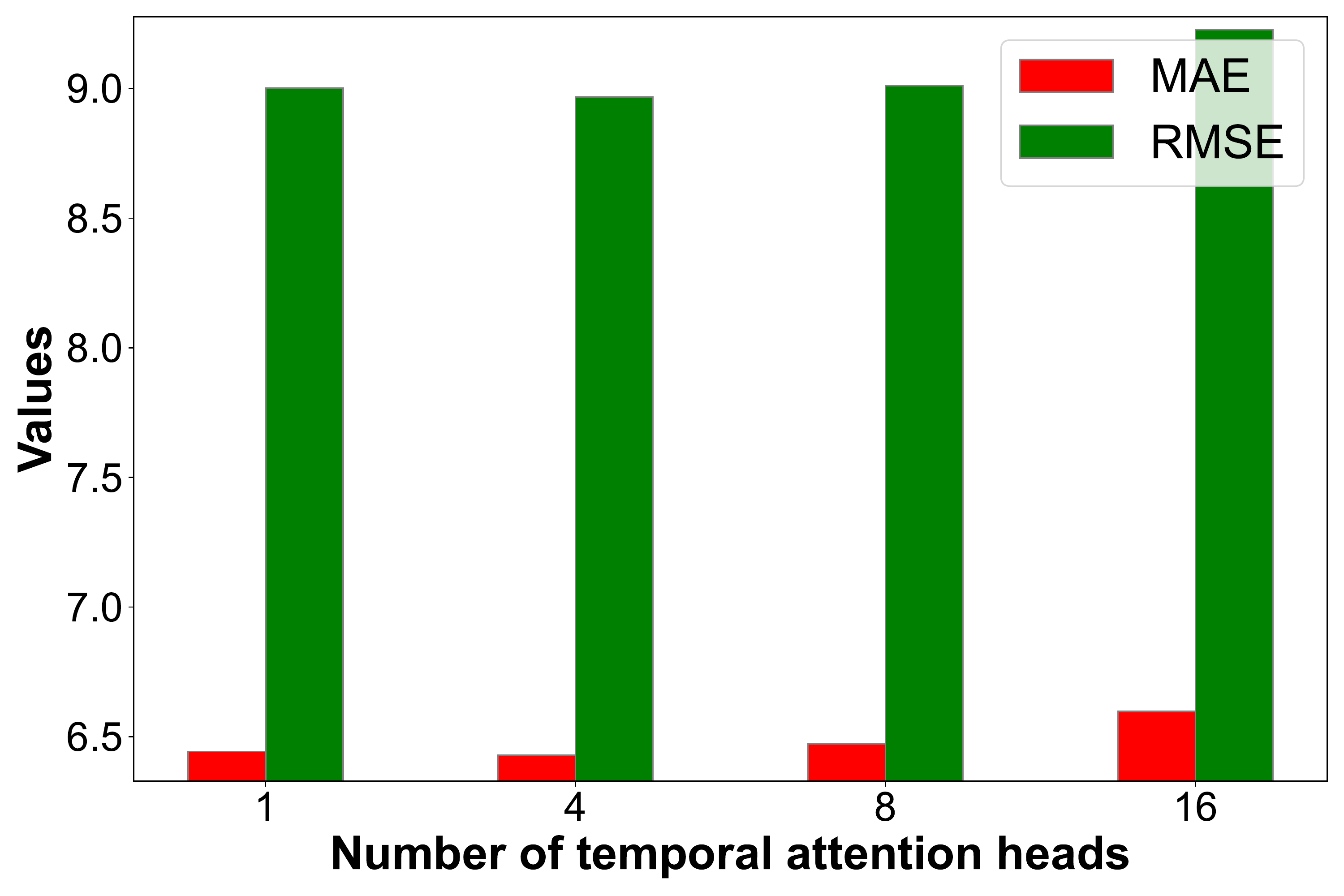}
    \caption{Varying number of temporal attention heads for each layer of STPN\--- optimal = 4.}
    \label{fig:head}
\end{figure}

\noindent \textbf{Number of temporal attention heads}: Assuming that the delay of multiple airports contains complex temporal dynamics, we use multiple attention heads to infer temporal adjacency matrices. We explore STPN with different numbers of temporal attention heads to judge the contribution of multiple temporal adjacency matrices. Results on U.S. dataset are shown in Figure~\ref{fig:head}. All models in Figure~\ref{fig:head} contain four layers of space-time-separable multi-graph convolution. The results show that the optimal number of attention heads is 4, slightly better than the one that only learns one temporal adjacency matrix in each layer. However, more heads induce difficulty in training, leading to increased errors. In Figure~\ref{fig:head}, the models with 8 and 16 temporal adjacency matrices give higher errors than the ones with four heads. It should be noted that STPN learns different temporal attention matrices in different layers. Therefore the model with one head also learns four types of temporal dependencies in this ablative experiment.

\begin{figure}[!ht]
    \centering
    \includegraphics[width = 0.45\textwidth]{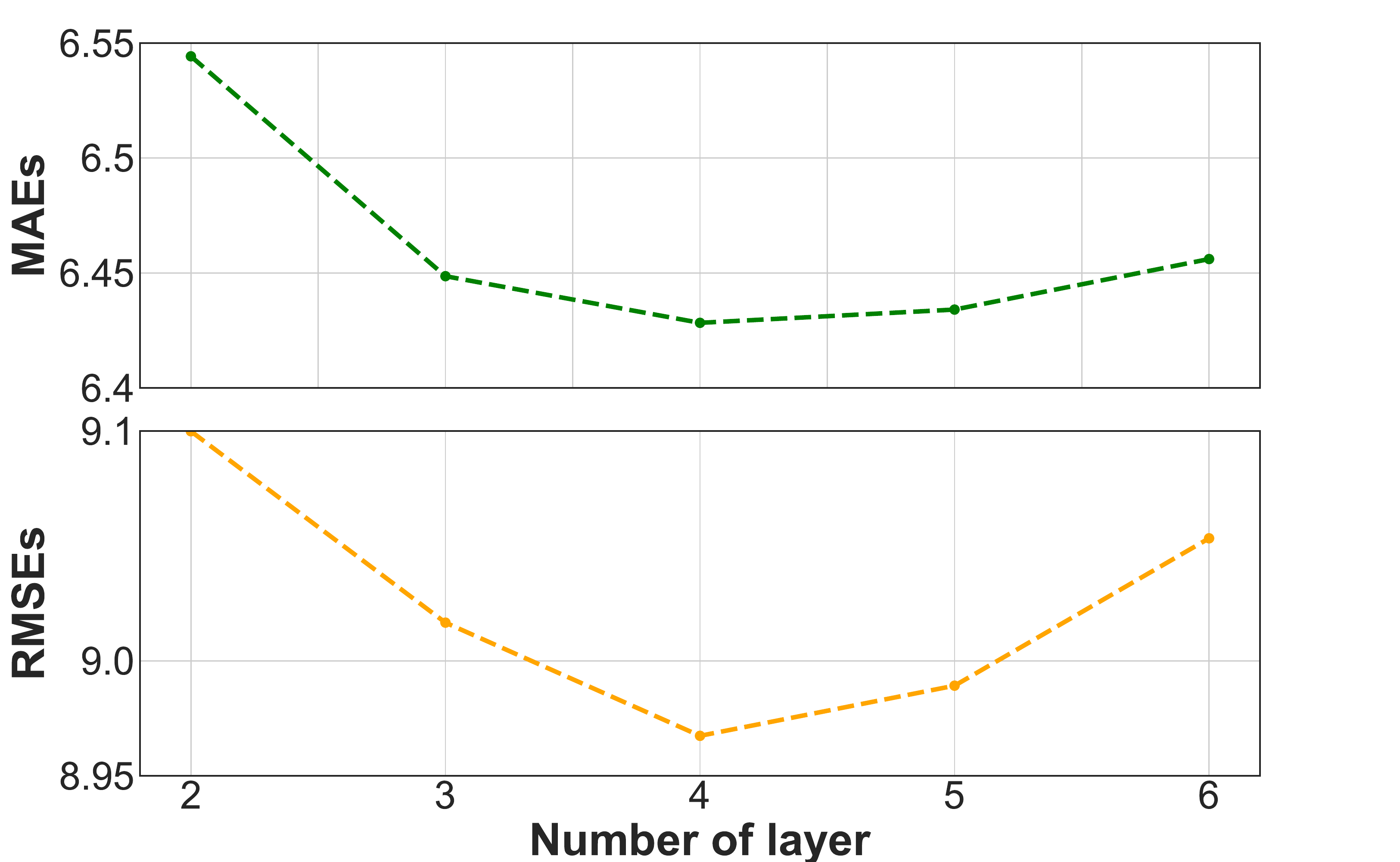}
    \caption{Varying number of Space-Time-Separable Multi-Graph Convolution layers for STPN\--- optimal = 4.}
    \label{fig:layer}
\end{figure}

\noindent \textbf{Number of layers}: The number of layers has a nonnegligible impact on the model's accuracy. With more layers, STPN can theoretically extract more complex spatiotemporal dependencies. However, training deeper graph convolutional networks is still challenging due to the over-smoothing issue \cite{li2018deeper, xu2018representation}.

\begin{figure*}[!htb]
\centering
\subfigure[Average departure delay of the historical inputs]{
\includegraphics[width = 0.3\textwidth]{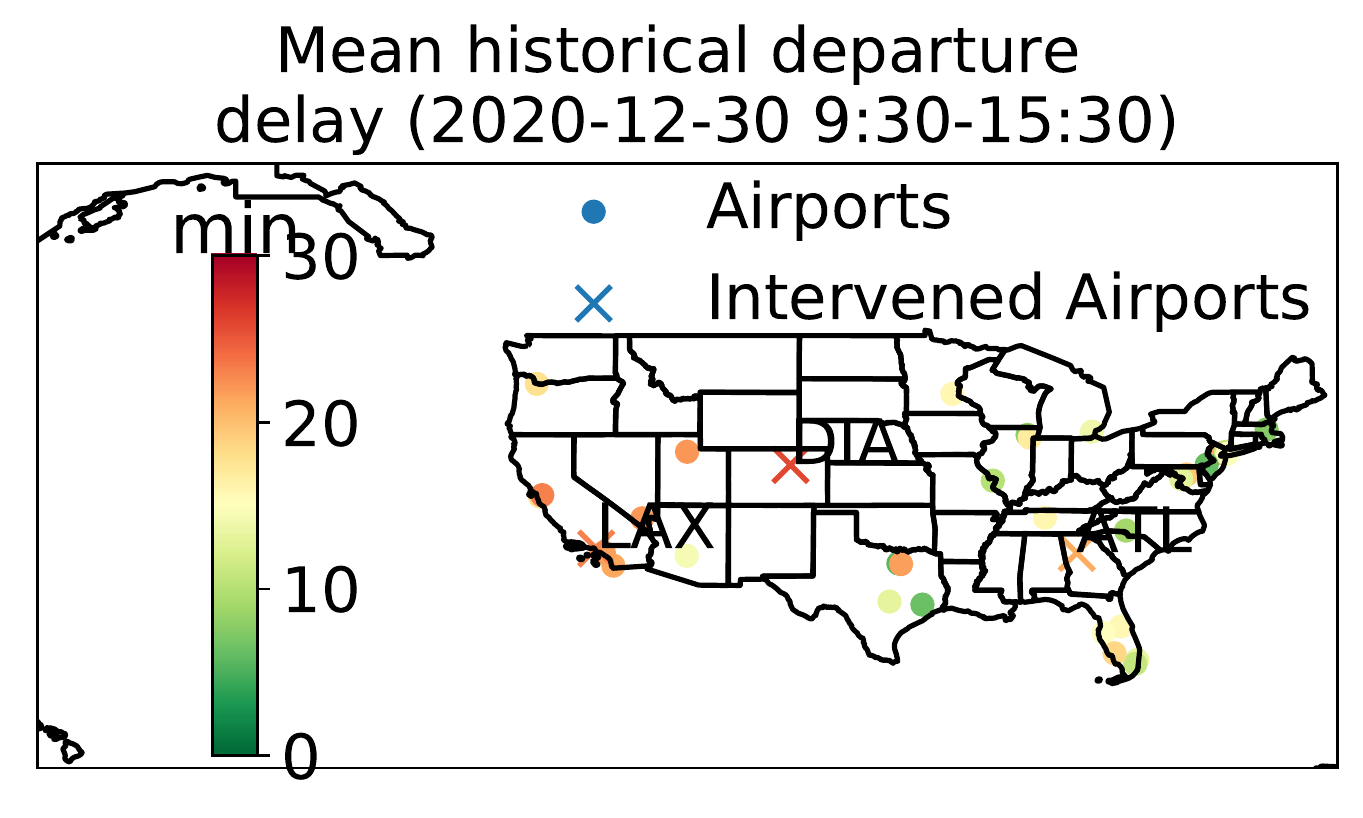}}
\subfigure[Predicted delay reduction of ATL]{
\includegraphics[width = 0.3\textwidth]{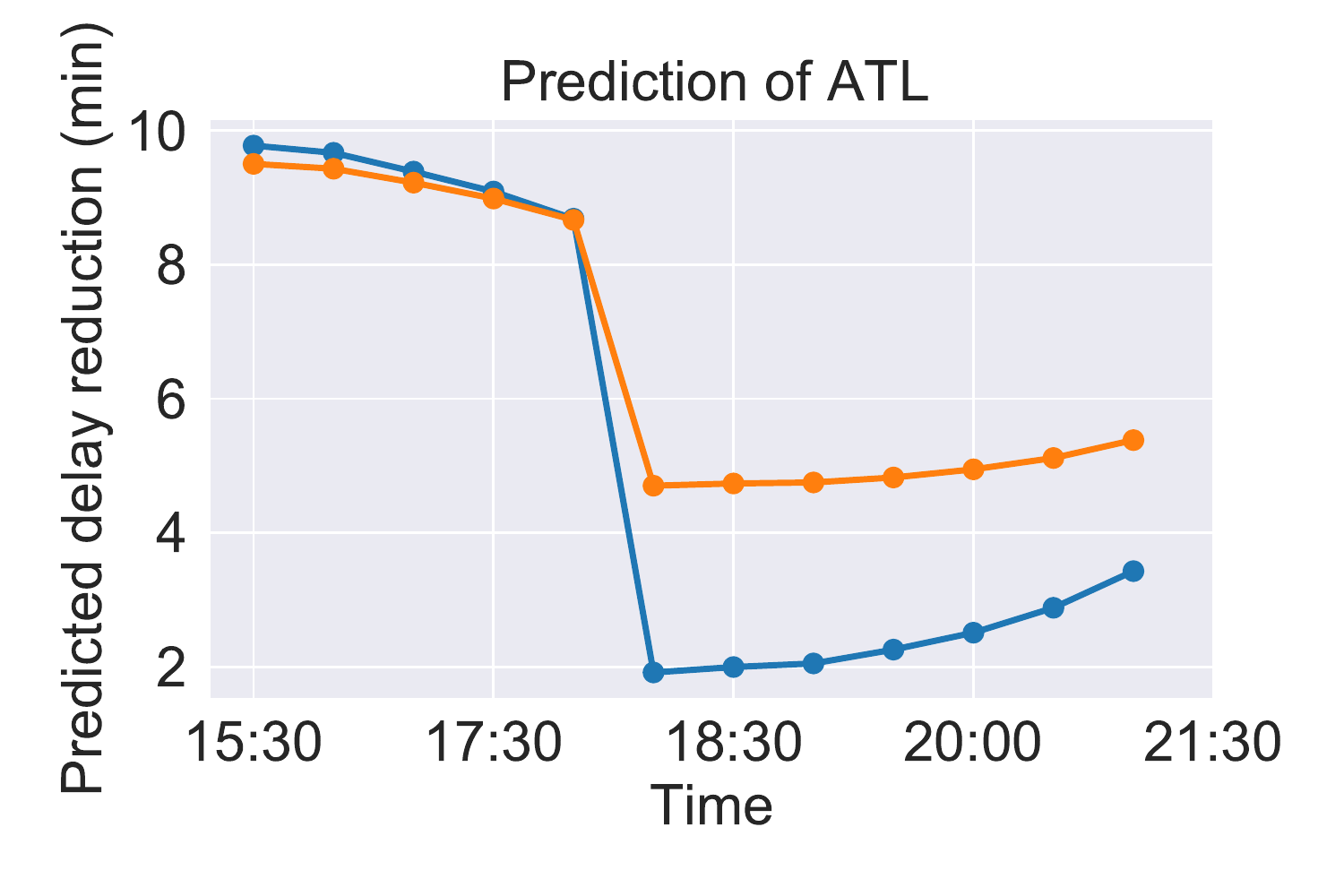}}
\subfigure[Predicted delay reduction of PDX]{
\includegraphics[width = 0.23\textwidth]{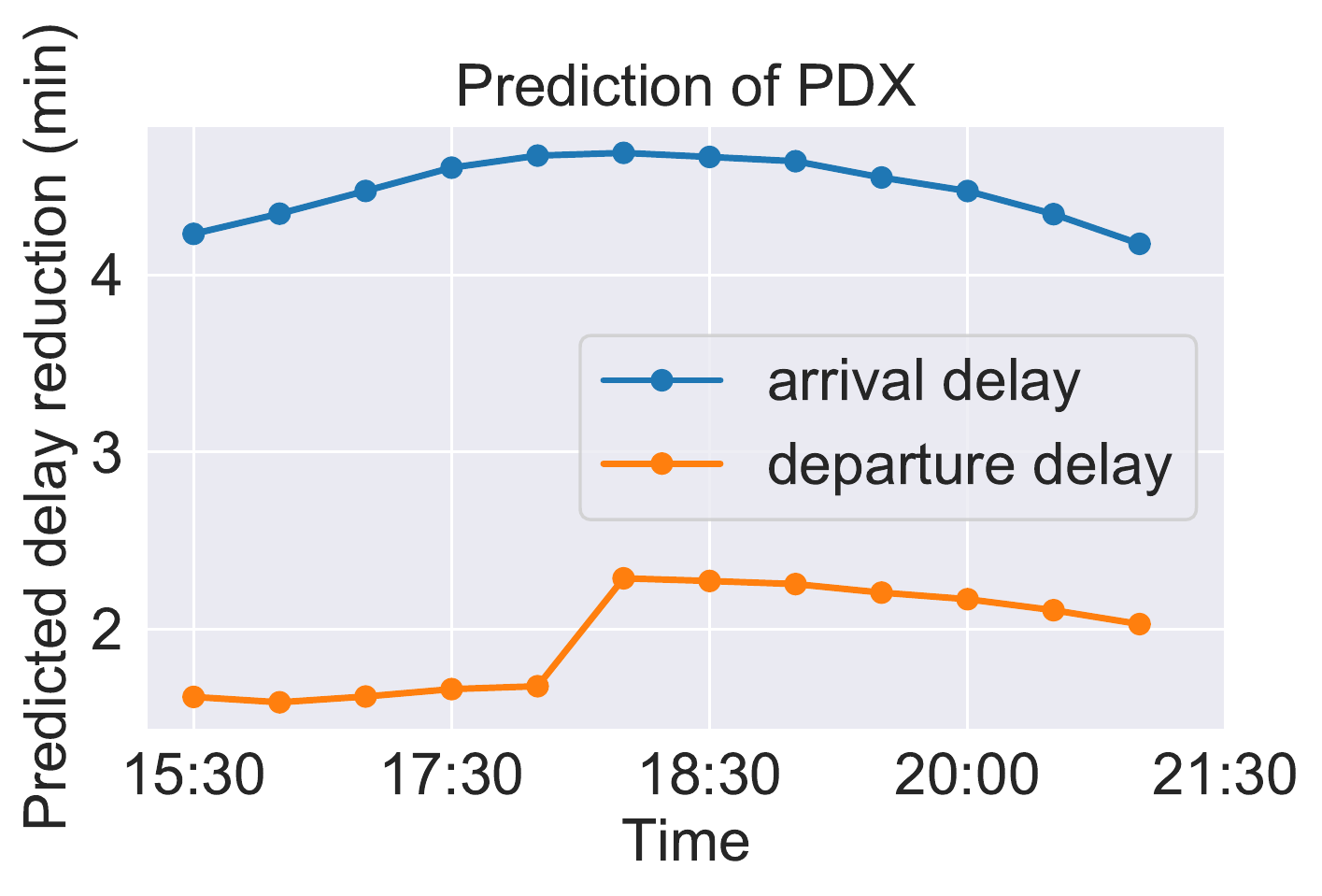}}
\subfigure[3 step ahead arrival delay reduction]{
\includegraphics[width = 0.3\textwidth]{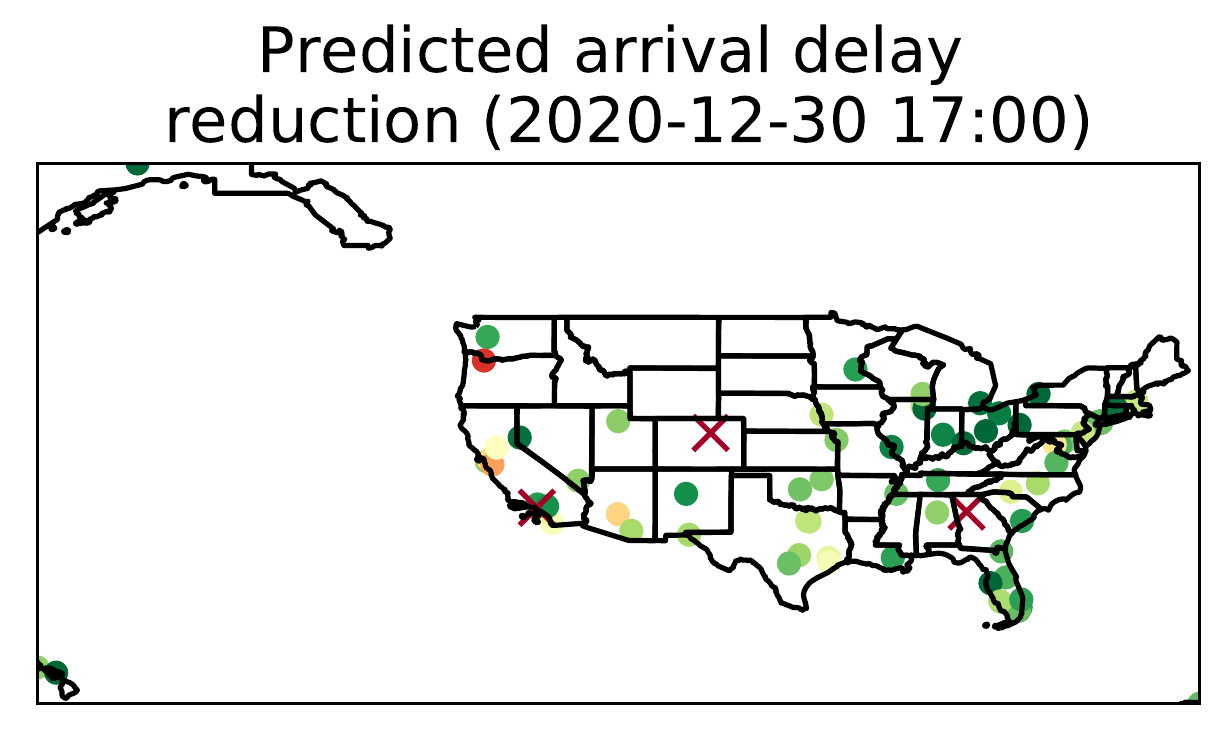}}
\subfigure[6 step ahead arrival delay reduction]{
\includegraphics[width = 0.3\textwidth]{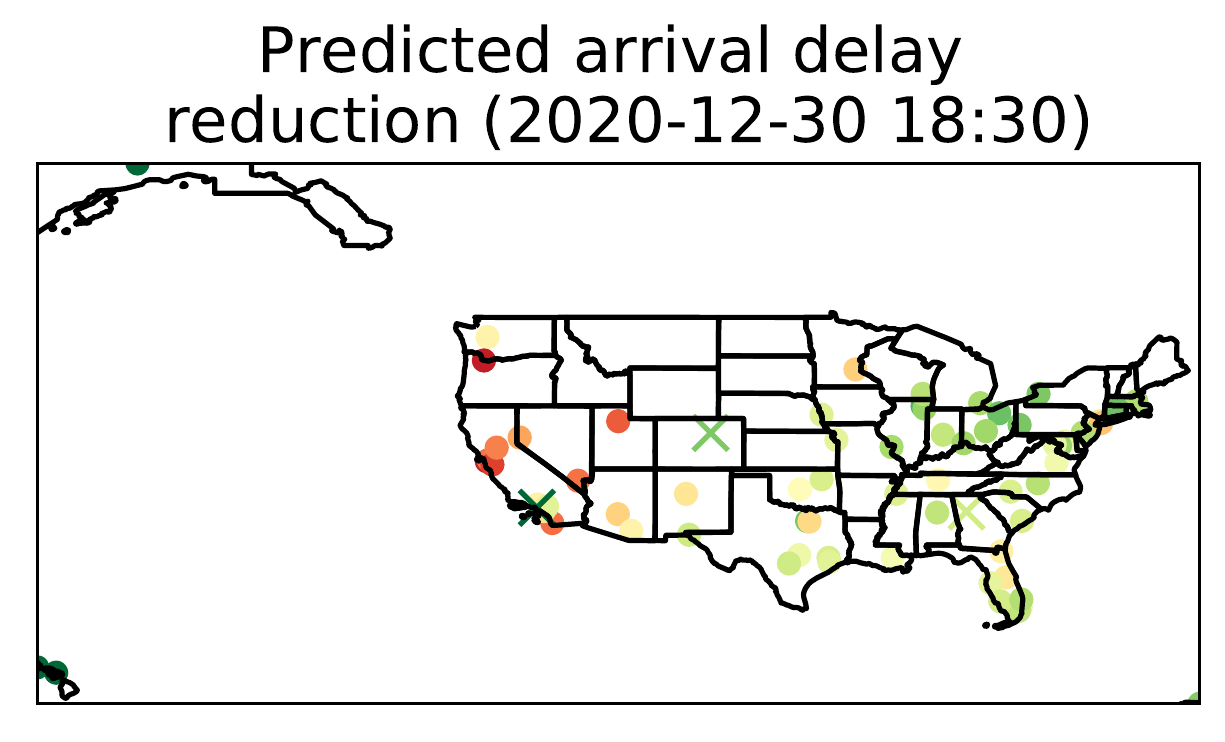}}
\subfigure[12 step ahead arrival delay reduction]{
\includegraphics[width = 0.3\textwidth]{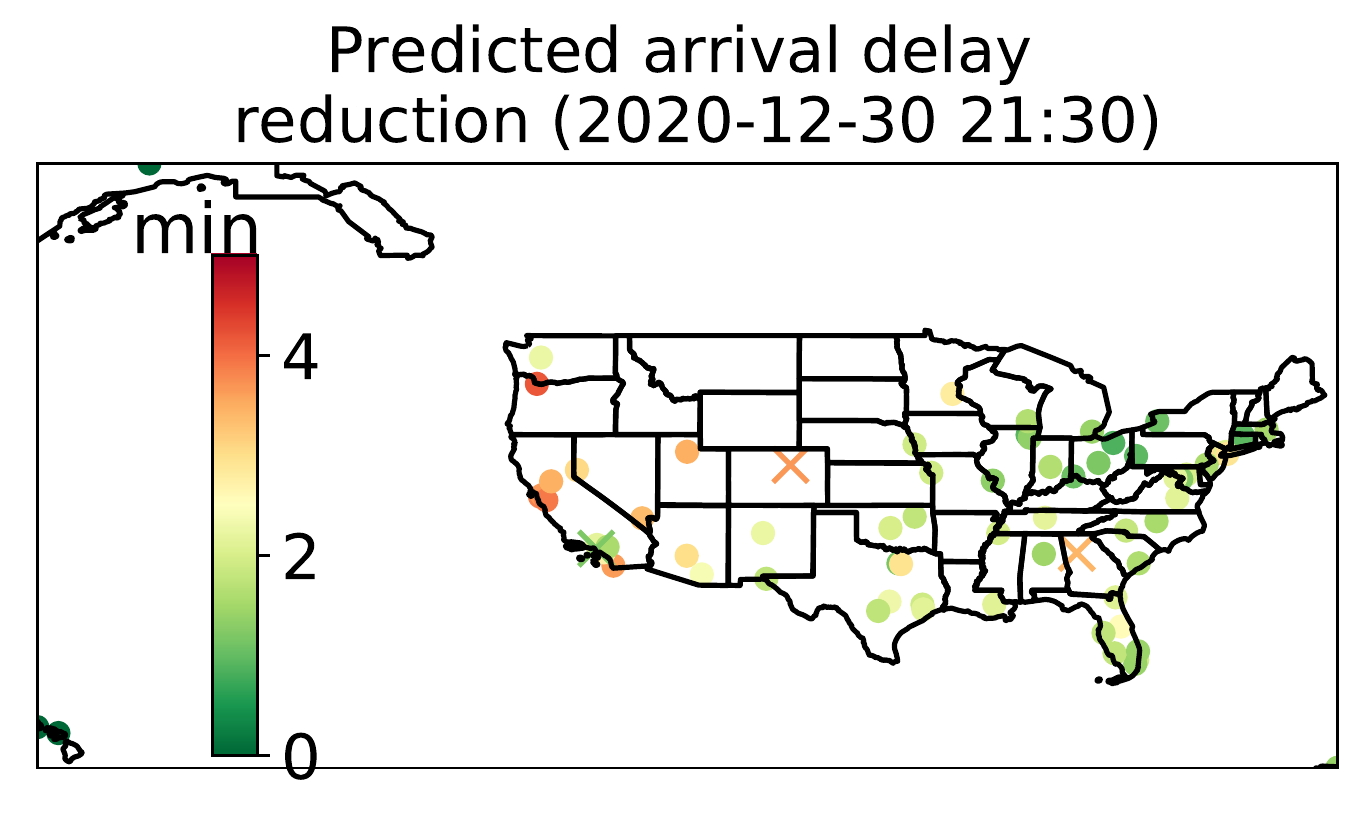}}
\subfigure[3 step ahead departure delay reduction]{
\includegraphics[width = 0.3\textwidth]{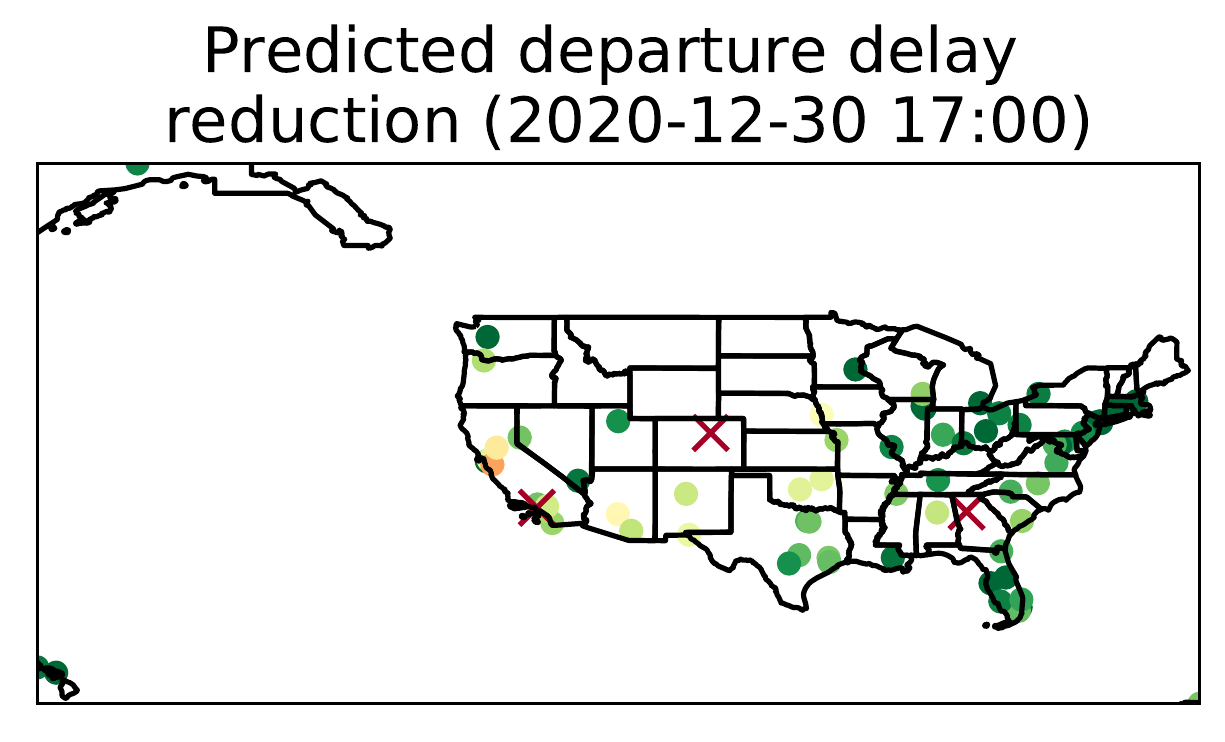}}
\subfigure[6 step ahead departure delay reduction]{
\includegraphics[width = 0.3\textwidth]{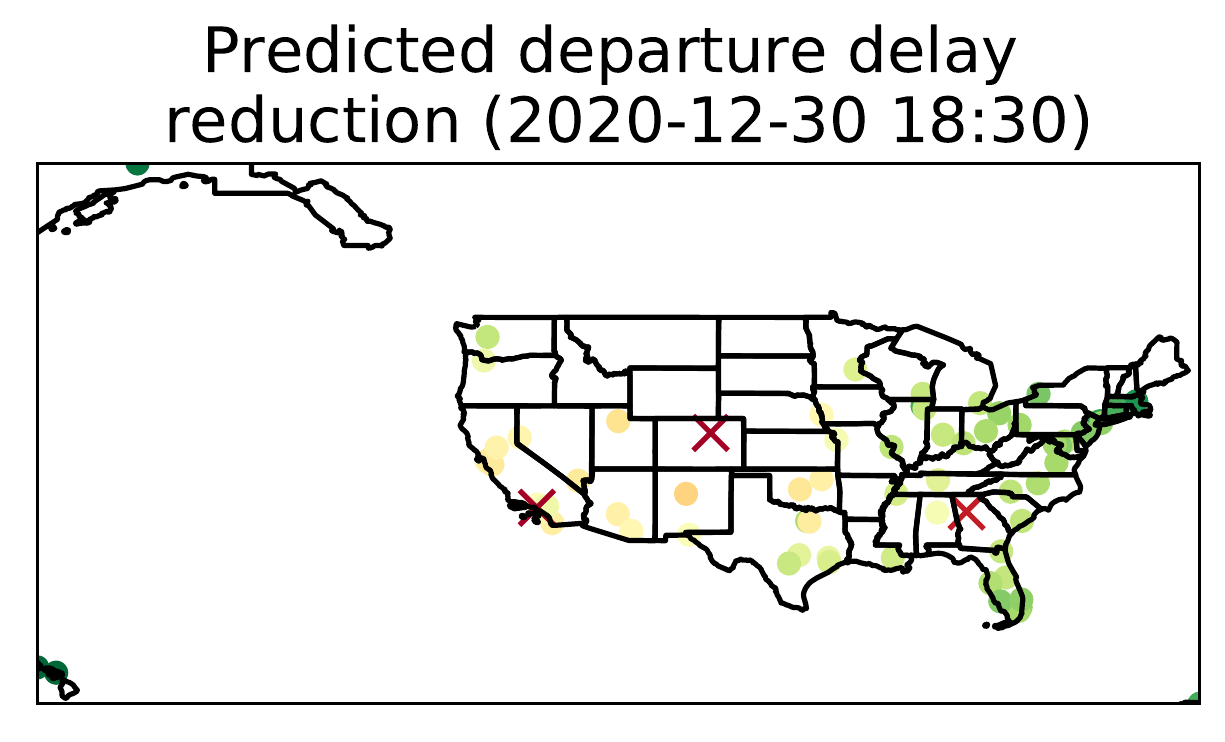}}
\subfigure[12 step ahead departure delay reduction]{
\includegraphics[width = 0.3\textwidth]{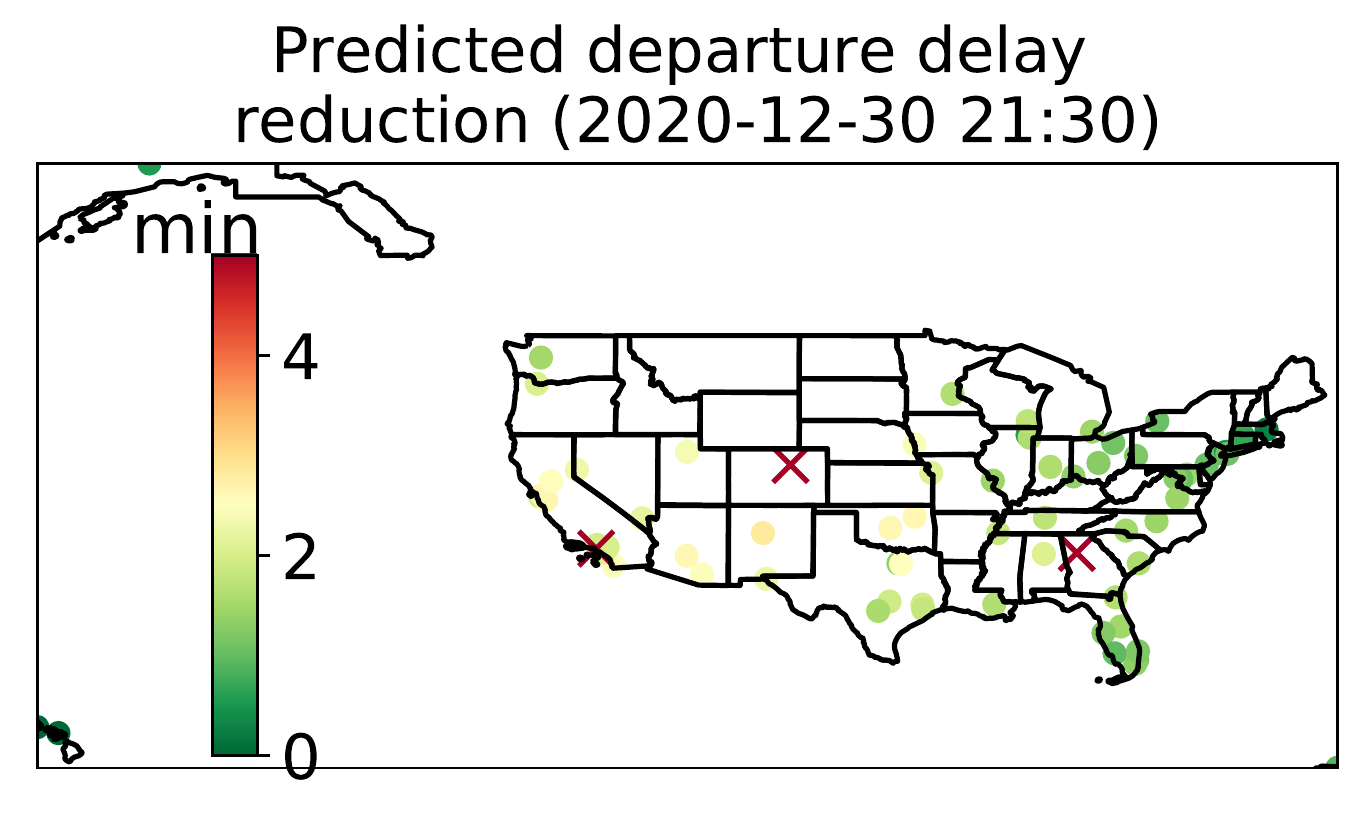}}
    \caption{Historical departure delay intervention results.}
    \label{fig:intervention}
\end{figure*}

We test two-layer STPN with 128 hidden neurons, three-layer STPN with [128, 64] hidden neurons, four-layer STPN with [128,64,32] hidden neurons, five-layer STPN with [128,64,32,32] hidden neurons, and six-layer STPN with [128,64,32,32,32] hidden neurons on U.S. dataset. Figure~\ref{fig:layer} shows that the four-layer model outperforms all other models. Our intuition is that this setting allows STPN to be trained easily due to its shallowness, and the four-layer model can extract more meaningful spatiotemporal delay propagation patterns than too shallow structures. However, we have also only performed this ablation study on a relatively shallow structure, so it might be the case that deeper architectures perform better. There is lots of room for future exploration in this design choice.

\subsection{Counterfactual Intervention}

The ultimate goal of the delay prediction model is to guide intervention for delay reduction. Therefore, the prediction model capable of guiding intervention should be able to think about the impact of alternatives to reality (counterfactuals) \cite{ijcai2019-876}. The counterfactual ability allows human users to simulate some aspects of the targeted system. Specifically, we are particularly interested in counterfactuals about future delay reduction. Reducing delay is a complex task that involves multiple controls of various degrees of granularity \cite{agogino2012multiagent}. In this work, we explore a specific intervention: we assume that we can reduce the historical departure delay of the busiest airports to 0s. This is reasonable since departure delays can be seen as the origin of all delays. After the intervention, we examine the delay reduction before and after intervention based on $\triangle x = x_{pred} - \hat{x}_{pred}$, in which $x_{pred}$ is the normal prediction with true input, and $\hat{x}_{pred}$ is the prediction with intervened inputs.

We depict an intervention example in Figure~\ref{fig:intervention}. In Figure~\ref{fig:intervention}, large-scale departure delays happened in the U.S. airport network. We reduce the historical departure delay of busy airports ATL, LAX, and DIA to 0s. The following observations should be highlighted from Figure~\ref{fig:intervention}: (1) The intervention dramatically reduces the departure and arrival delays of the intervened airport in a short-term temporal range. The arrival and departure delay reductions were larger than 8 minutes from 16:00 to 18:00. As expected, the intervention, which reduces the departure delay, can increase the capacity of airports. As a result, it can dramatically reduce future arrival and departure delays. After a short period (18:00 in Figure~\ref{fig:intervention}(a-c)), the intervention effects dropped. The reason might be that the flight delay from other airports will eventually cascade to ATL and mitigate the effects of the intervention. (2) The intervention can also dramatically reduce the arrival delays of parts airports. Figure~\ref{fig:intervention}(c) gives an example of PDX airport, whose arrival delay has been significantly influenced by the intervention. The arrival delay is dropped by 4 minutes, while the departure delay of PDX has not been greatly affected. It means that if an aircraft reduces its departure delay, this reduced departure delay may directly reduce the arrival delay, and it can indirectly propagates to departure delay. We also observe the reduction of arrival delay drops, and the one of departure delay rises after 18:00. The shortest travel time from DIA and LAX to PDX ranges from 2 to 2.5 hours. After 2.5 hours, only the intervened flights from ATL could directly impact arrival delay. The results show that STPN has learned this physical relationship. (3) The local delay reduction can propagate through part or all of the airport network. In Figure~\ref{fig:intervention}(d-f), the intervention nearly causes a great delay reduction in all Northwestern airports. After 6 hours, we also observed a large reduction of arrival delays in the intervened airports. The more times the intervened aircraft takes off from intervened airports, the more susceptible it becomes to upstream delay reduction that may affect subsequent visits to intervened airports. In a nutshell, reduced departure delay of the intervened airport causes high arrival delay reduction onto the intervened airport itself. Compared with arrival delay, the intervention only has local effects on departure delay. In Figure~\ref{fig:intervention}(g-i), the reductions of departure delays are far smaller than those of arrival delays in Figure~\ref{fig:intervention}(d-f), except for the intervened airports. In summary, those counterfactual observations show that our STPN has learned some realistic delay propagation patterns from the dataset. We aim to investigate this subject more carefully in future work.

\section{Conclusion and Future Works}
\label{sec:con}

We have presented a deep learning model named STPN for predicting multi-step arrival and departure delay in large networks of airports, taking into account the spatiotemporal dependencies, exogenous factor effects, and departure-arrival delay relationship. Because of the efficiencies gained through the space-time-separable multi-graph convolution framework we have adopted, as well as several specific structure designs for spatiotemporal dependencies mining, the STPN provides more accurate predictions compared with several baselines. Moreover, the counterfactuals produced by STPN can be a powerful tool for exploring, from a macroscopic perspective, the implications of a broad range of alternative strategies concerning delay reduction interventions in a regional or national system of airports. 

Delay propagation modeling is a central notion in modern air traffic management systems, which aims to exploit the compact spatiotemporal dependencies underlying real-world aviation systems. Granted, the possibilities for future research from this paper are extensive. From the perspective of deep neural networks for delay propagation learning, ongoing work involves more advanced structure design. A dynamic extension of this work's static sum-of-Kronecker spatiotemporal kernel would be desirable. Further extensions could include the structure for handling missing data. It would also be interesting to explore how effectively the model can be used for small airports with sparse flights. In the aspects of real-world applications, the STPN only works from the macroscopic perspective. Developing models for learning delay propagation between different individual aircraft at the mesoscopic level would also be necessary. To that end, the work presented in this paper suggests new insights on data-driven delay propagation learning that may benefit air traffic management.

\section*{Acknowledgement}

This work was supported by National Natural Science Foundation of China (Grant No.U20A20161), the Open Fund of Key Laboratory of Flight Techniques and Flight Safety, Civil Aviation Administration of China (Grant No. FZ2021KF04), and the Fundamental Research Funds for the Central Universities.

\bibliographystyle{IEEEtran}
\bibliography{mgcnn}

\begin{thebibliography}{10}
\providecommand{\url}[1]{#1}
\csname url@samestyle\endcsname
\providecommand{\newblock}{\relax}
\providecommand{\bibinfo}[2]{#2}
\providecommand{\BIBentrySTDinterwordspacing}{\spaceskip=0pt\relax}
\providecommand{\BIBentryALTinterwordstretchfactor}{4}
\providecommand{\BIBentryALTinterwordspacing}{\spaceskip=\fontdimen2\font plus
\BIBentryALTinterwordstretchfactor\fontdimen3\font minus
  \fontdimen4\font\relax}
\providecommand{\BIBforeignlanguage}[2]{{%
\expandafter\ifx\csname l@#1\endcsname\relax
\typeout{** WARNING: IEEEtran.bst: No hyphenation pattern has been}%
\typeout{** loaded for the language `#1'. Using the pattern for}%
\typeout{** the default language instead.}%
\else
\language=\csname l@#1\endcsname
\fi
#2}}
\providecommand{\BIBdecl}{\relax}
\BIBdecl

\bibitem{IATA}
``Iata economics,'' International Air Transport Association (IATA), Tech. Rep.,
  2020.

\bibitem{airport_cost}
N.~de~Loisy, \emph{Transportation and the Belt and Road Initiative}.\hskip 1em
  plus 0.5em minus 0.4em\relax Supply Chain Management Outsource Limited, 2019.

\bibitem{bertsimas2011integer}
D.~Bertsimas, G.~Lulli, and A.~Odoni, ``An integer optimization approach to
  large-scale air traffic flow management,'' \emph{Operations research},
  vol.~59, no.~1, pp. 211--227, 2011.

\bibitem{bertsimas2016fairness}
D.~Bertsimas and S.~Gupta, ``Fairness and collaboration in network air traffic
  flow management: an optimization approach,'' \emph{Transportation Science},
  vol.~50, no.~1, pp. 57--76, 2016.

\bibitem{ayhan2016aircraft}
S.~Ayhan and H.~Samet, ``Aircraft trajectory prediction made easy with
  predictive analytics,'' in \emph{Proceedings of the 22nd ACM SIGKDD
  International Conference on Knowledge Discovery and Data Mining}, 2016, pp.
  21--30.

\bibitem{kafle2016modeling}
N.~Kafle and B.~Zou, ``Modeling flight delay propagation: A new
  analytical-econometric approach,'' \emph{Transportation Research Part B:
  Methodological}, vol.~93, pp. 520--542, 2016.

\bibitem{li2021graph}
M.~Z. Li, K.~Gopalakrishnan, K.~Pantoja, and H.~Balakrishnan, ``Graph signal
  processing techniques for analyzing aviation disruptions,''
  \emph{Transportation Science}, vol.~55, no.~3, pp. 553--573, 2021.

\bibitem{cai2021spatial}
Q.~Cai, S.~Alam, and V.~N. Duong, ``A spatial--temporal network perspective for
  the propagation dynamics of air traffic delays,'' \emph{Engineering}, vol.~7,
  no.~4, pp. 452--464, 2021.

\bibitem{pyrgiotis2013modelling}
N.~Pyrgiotis, K.~M. Malone, and A.~Odoni, ``Modelling delay propagation within
  an airport network,'' \emph{Transportation Research Part C: Emerging
  Technologies}, vol.~27, pp. 60--75, 2013.

\bibitem{simaiakis2016queuing}
I.~Simaiakis and H.~Balakrishnan, ``A queuing model of the airport departure
  process,'' \emph{Transportation Science}, vol.~50, no.~1, pp. 94--109, 2016.

\bibitem{rebollo2014characterization}
J.~J. Rebollo and H.~Balakrishnan, ``Characterization and prediction of air
  traffic delays,'' \emph{Transportation research part C: Emerging
  technologies}, vol.~44, pp. 231--241, 2014.

\bibitem{yu2019flight}
B.~Yu, Z.~Guo, S.~Asian, H.~Wang, and G.~Chen, ``Flight delay prediction for
  commercial air transport: A deep learning approach,'' \emph{Transportation
  Research Part E: Logistics and Transportation Review}, vol. 125, pp.
  203--221, 2019.

\bibitem{bao2021graph}
J.~Bao, Z.~Yang, and W.~Zeng, ``Graph to sequence learning with attention
  mechanism for network-wide multi-step-ahead flight delay prediction,''
  \emph{Transportation Research Part C: Emerging Technologies}, vol. 130, p.
  103323, 2021.

\bibitem{cai2021deep}
K.~Cai, Y.~Li, Y.-P. Fang, and Y.~Zhu, ``A deep learning approach for flight
  delay prediction through time-evolving graphs,'' \emph{IEEE Transactions on
  Intelligent Transportation Systems}, 2021.

\bibitem{sofianos2021space}
T.~Sofianos, A.~Sampieri, L.~Franco, and F.~Galasso, ``Space-time-separable
  graph convolutional network for pose forecasting,'' in \emph{Proceedings of
  the IEEE/CVF International Conference on Computer Vision}, 2021, pp.
  11\,209--11\,218.

\bibitem{vaswani2017attention}
A.~Vaswani, N.~Shazeer, N.~Parmar, J.~Uszkoreit, L.~Jones, A.~N. Gomez,
  {\L}.~Kaiser, and I.~Polosukhin, ``Attention is all you need,''
  \emph{Advances in neural information processing systems}, vol.~30, 2017.

\bibitem{li2018diffusion}
Y.~Li, R.~Yu, C.~Shahabi, and Y.~Liu, ``Diffusion convolutional recurrent
  neural network: Data-driven traffic forecasting,'' in \emph{International
  Conference on Learning Representations}, 2018.

\bibitem{hu2018squeeze}
J.~Hu, L.~Shen, and G.~Sun, ``Squeeze-and-excitation networks,'' in
  \emph{Proceedings of the IEEE conference on computer vision and pattern
  recognition}, 2018, pp. 7132--7141.

\bibitem{beatty1999preliminary}
R.~Beatty, R.~Hsu, L.~Berry, and J.~Rome, ``Preliminary evaluation of flight
  delay propagation through an airline schedule,'' \emph{Air Traffic Control
  Quarterly}, vol.~7, no.~4, pp. 259--270, 1999.

\bibitem{fleurquin2013systemic}
P.~Fleurquin, J.~J. Ramasco, and V.~M. Eguiluz, ``Systemic delay propagation in
  the us airport network,'' \emph{Scientific reports}, vol.~3, no.~1, pp. 1--6,
  2013.

\bibitem{campanelli2016comparing}
B.~Campanelli, P.~Fleurquin, A.~Arranz, I.~Etxebarria, C.~Ciruelos, V.~M.
  Egu{\'\i}luz, and J.~J. Ramasco, ``Comparing the modeling of delay
  propagation in the us and european air traffic networks,'' \emph{Journal of
  Air Transport Management}, vol.~56, pp. 12--18, 2016.

\bibitem{wu2019modelling}
C.-L. Wu and K.~Law, ``Modelling the delay propagation effects of multiple
  resource connections in an airline network using a bayesian network model,''
  \emph{Transportation Research Part E: Logistics and Transportation Review},
  vol. 122, pp. 62--77, 2019.

\bibitem{nayak2011estimation}
N.~Nayak and Y.~Zhang, ``Estimation and comparison of impact of single airport
  delay on national airspace system with multivariate simultaneous models,''
  \emph{Transportation research record}, vol. 2206, no.~1, pp. 52--60, 2011.

\bibitem{wu2019comprehensive}
Z.~Wu, S.~Pan, F.~Chen, G.~Long, C.~Zhang, and P.~S. Yu, ``A comprehensive
  survey on graph neural networks,'' \emph{arXiv preprint arXiv:1901.00596},
  2019.

\bibitem{bruna2014spectral}
J.~Bruna, W.~Zaremba, A.~Szlam, and Y.~LeCun, ``Spectral networks and deep
  locally connected networks on graphs,'' in \emph{2nd International Conference
  on Learning Representations, ICLR 2014}, 2014.

\bibitem{kipf2016semi}
T.~N. Kipf and M.~Welling, ``Semi-supervised classification with graph
  convolutional networks,'' \emph{arXiv preprint arXiv:1609.02907}, 2016.

\bibitem{defferrard2016convolutional}
M.~Defferrard, X.~Bresson, and P.~Vandergheynst, ``Convolutional neural
  networks on graphs with fast localized spectral filtering,'' in
  \emph{Advances in neural information processing systems}, 2016, pp.
  3844--3852.

\bibitem{atwood2016diffusion}
J.~Atwood and D.~Towsley, ``Diffusion-convolutional neural networks,''
  \emph{Advances in neural information processing systems}, vol.~29, 2016.

\bibitem{zhu2020simple}
H.~Zhu and P.~Koniusz, ``Simple spectral graph convolution,'' in
  \emph{International Conference on Learning Representations}, 2020.

\bibitem{seo2018structured}
Y.~Seo, M.~Defferrard, P.~Vandergheynst, and X.~Bresson, ``Structured sequence
  modeling with graph convolutional recurrent networks,'' in
  \emph{International Conference on Neural Information Processing}.\hskip 1em
  plus 0.5em minus 0.4em\relax Springer, 2018, pp. 362--373.

\bibitem{yu2018spatio}
B.~Yu, H.~Yin, and Z.~Zhu, ``Spatio-temporal graph convolutional networks: a
  deep learning framework for traffic forecasting,'' in \emph{Proceedings of
  the 27th International Joint Conference on Artificial Intelligence}.\hskip
  1em plus 0.5em minus 0.4em\relax AAAI Press, 2018, pp. 3634--3640.

\bibitem{wu2019graph}
Z.~Wu, S.~Pan, G.~Long, J.~Jiang, and C.~Zhang, ``Graph wavenet for deep
  spatial-temporal graph modeling,'' in \emph{Proceedings of the 28th
  International Joint Conference on Artificial Intelligence}, 2019, pp.
  1907--1913.

\bibitem{zheng2020gman}
C.~Zheng, X.~Fan, C.~Wang, and J.~Qi, ``Gman: A graph multi-attention network
  for traffic prediction,'' in \emph{Proceedings of the AAAI Conference on
  Artificial Intelligence}, vol.~34, no.~01, 2020, pp. 1234--1241.

\bibitem{guo2021learning}
S.~Guo, Y.~Lin, H.~Wan, X.~Li, and G.~Cong, ``Learning dynamics and
  heterogeneity of spatial-temporal graph data for traffic forecasting,''
  \emph{IEEE Transactions on Knowledge and Data Engineering}, 2021.

\bibitem{monti2017geometric}
F.~Monti, M.~Bronstein, and X.~Bresson, ``Geometric matrix completion with
  recurrent multi-graph neural networks,'' in \emph{Advances in Neural
  Information Processing Systems}, 2017, pp. 3697--3707.

\bibitem{kolda2009tensor}
T.~G. Kolda and B.~W. Bader, ``Tensor decompositions and applications,''
  \emph{SIAM review}, vol.~51, no.~3, pp. 455--500, 2009.

\bibitem{bonilla2007multi}
E.~V. Bonilla, K.~Chai, and C.~Williams, ``Multi-task gaussian process
  prediction,'' \emph{Advances in neural information processing systems},
  vol.~20, 2007.

\bibitem{lan2019learning}
S.~Lan, ``Learning temporal evolution of spatial dependence with generalized
  spatiotemporal gaussian process models,'' \emph{arXiv preprint
  arXiv:1901.04030}, 2019.

\bibitem{wu2021inductive}
Y.~Wu, D.~Zhuang, A.~Labbe, and L.~Sun, ``Inductive graph neural networks for
  spatiotemporal kriging,'' in \emph{Proceedings of the AAAI Conference on
  Artificial Intelligence}, vol.~35, no.~5, 2021, pp. 4478--4485.

\bibitem{wang2020position}
B.~Wang, L.~Shang, C.~Lioma, X.~Jiang, H.~Yang, Q.~Liu, and J.~G. Simonsen,
  ``On position embeddings in bert,'' in \emph{International Conference on
  Learning Representations}, 2020.

\bibitem{wilson2014fast}
A.~G. Wilson, E.~Gilboa, A.~Nehorai, and J.~P. Cunningham, ``Fast kernel
  learning for multidimensional pattern extrapolation,'' \emph{Advances in
  Neural Information Processing Systems}, vol.~27, 2014.

\bibitem{moosavi2019short}
S.~Moosavi, M.~H. Samavatian, A.~Nandi, S.~Parthasarathy, and R.~Ramnath,
  ``Short and long-term pattern discovery over large-scale geo-spatiotemporal
  data,'' in \emph{Proceedings of the 25th ACM SIGKDD International Conference
  on Knowledge Discovery \& Data Mining}, 2019, pp. 2905--2913.

\bibitem{menon2004new}
P.~K. Menon, G.~D. Sweriduk, and K.~D. Bilimoria, ``New approach for modeling,
  analysis, and control of air traffic flow,'' \emph{Journal of guidance,
  control, and dynamics}, vol.~27, no.~5, pp. 737--744, 2004.

\bibitem{jain2019attention}
S.~Jain and B.~C. Wallace, ``Attention is not explanation,'' in
  \emph{Proceedings of the 2019 Conference of the North American Chapter of the
  Association for Computational Linguistics: Human Language Technologies,
  Volume 1 (Long and Short Papers)}, 2019, pp. 3543--3556.

\bibitem{li2019enhancing}
S.~Li, X.~Jin, Y.~Xuan, X.~Zhou, W.~Chen, Y.-X. Wang, and X.~Yan, ``Enhancing
  the locality and breaking the memory bottleneck of transformer on time series
  forecasting,'' \emph{Advances in neural information processing systems},
  vol.~32, 2019.

\bibitem{chefer2021transformer}
H.~Chefer, S.~Gur, and L.~Wolf, ``Transformer interpretability beyond attention
  visualization,'' in \emph{Proceedings of the IEEE/CVF Conference on Computer
  Vision and Pattern Recognition}, 2021, pp. 782--791.

\bibitem{li2018deeper}
Q.~Li, Z.~Han, and X.-M. Wu, ``Deeper insights into graph convolutional
  networks for semi-supervised learning,'' in \emph{Thirty-Second AAAI
  Conference on Artificial Intelligence}, 2018.

\bibitem{xu2018representation}
K.~Xu, C.~Li, Y.~Tian, T.~Sonobe, K.-i. Kawarabayashi, and S.~Jegelka,
  ``Representation learning on graphs with jumping knowledge networks,'' in
  \emph{International conference on machine learning}.\hskip 1em plus 0.5em
  minus 0.4em\relax PMLR, 2018, pp. 5453--5462.

\bibitem{ijcai2019-876}
R.~M.~J. Byrne, ``Counterfactuals in explainable artificial intelligence (xai):
  Evidence from human reasoning,'' in \emph{Proceedings of the Twenty-Eighth
  International Joint Conference on Artificial Intelligence, {IJCAI-19}}, 7
  2019, pp. 6276--6282.

\bibitem{agogino2012multiagent}
A.~K. Agogino and K.~Tumer, ``A multiagent approach to managing air traffic
  flow,'' \emph{Autonomous Agents and Multi-Agent Systems}, vol.~24, no.~1, pp.
  1--25, 2012.

\end{thebibliography}

\begin{IEEEbiography}[{\includegraphics[width=1in,height=1.25in,clip,keepaspectratio]{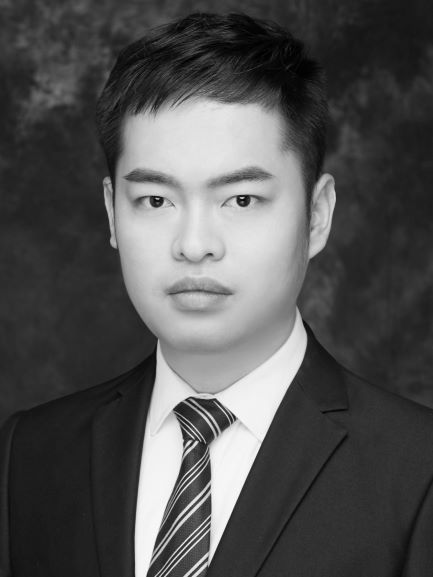}}]
{Yuankai Wu} (Member, IEEE) received the M.S. degree in transportation engineering and the Ph.D. degree in vehicle operation engineering, both from the Beijing Institute of Technology (BIT), Beijing, China, in 2015 and 2019. He is a full professor at the College of Computer Science, Sichuan University, China. Prior to joining Sichuan University in March 2022, he was an IVADO postdoc researcher with the Department of Civil Engineering, McGill University. His research interests include spatiotemporal data analysis, intelligent transportation systems, and intelligent decision-making.
\end{IEEEbiography}

\begin{IEEEbiography}[{\includegraphics[width=1in,height=1.25in,clip,keepaspectratio]{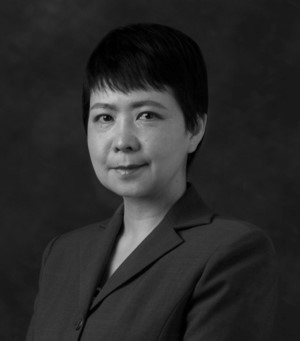}}]
{Hongyu Yang} received B.S. degree from Shanghai Jiaotong University in 1998, and received M.S. and Ph.D. degree from Sichuan University in 1991 and 2008 respectively. She is currently a professor with the College of Computer Science, Sichuan University, China. Her research interests include visual synthesis, digital image processing, computer graphics, real-time software engineering, air traffic information intelligent processing technology, visual navigation, and flight simulation technology.
\end{IEEEbiography}

\begin{IEEEbiography}[{\includegraphics[width=1in,height=1.25in,clip,keepaspectratio]{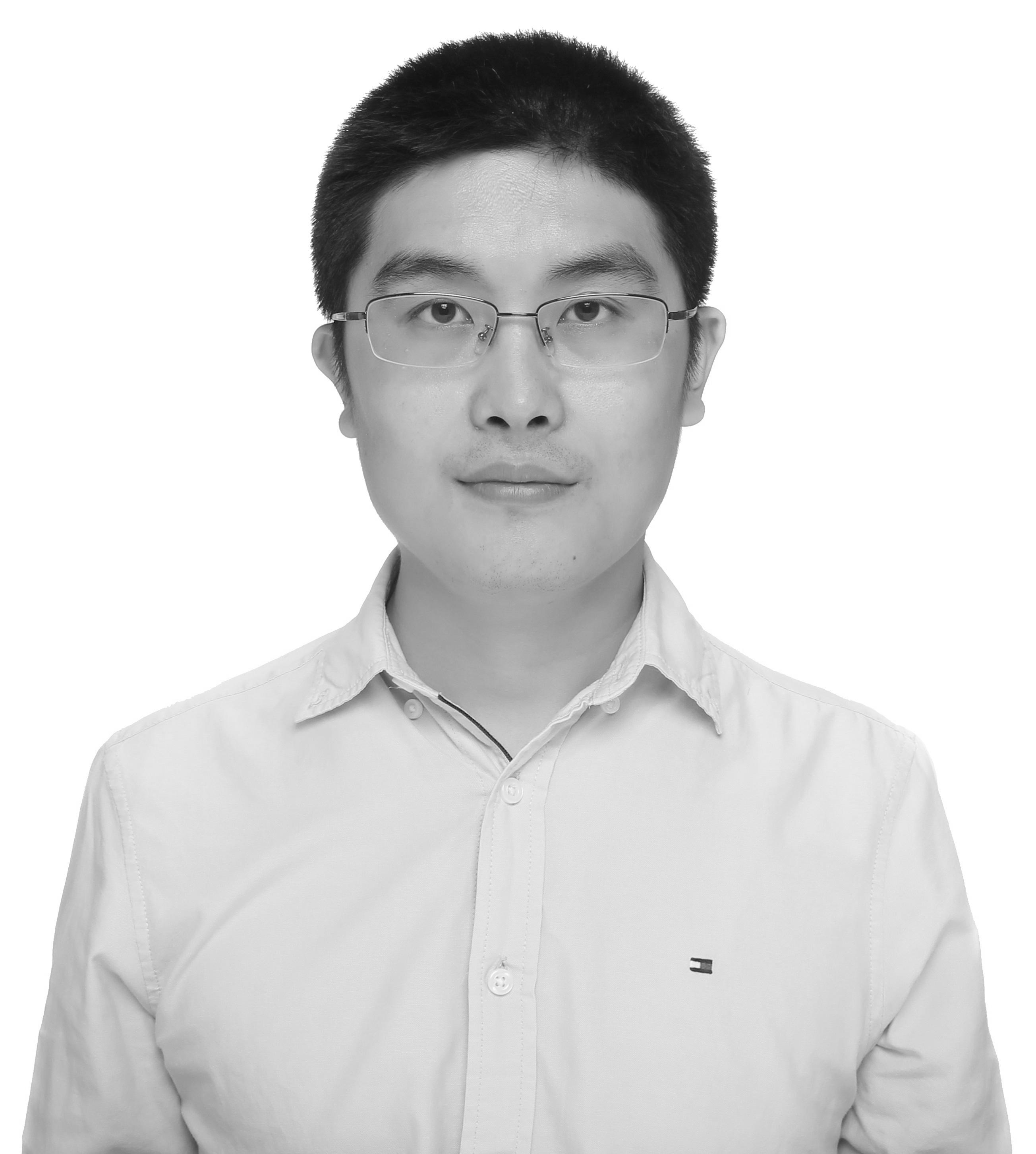}}]
{Yi Lin} (Member, IEEE) received the Ph.D. degree from Sichuan University, Chengdu, China, in 2019. He currently works as an Associate Professor with the College of Computer Science, Sichuan University. He was also a Visiting Scholar with the Department of Civil and Environmental Engineering, University of Wisconsin–Madison, Madison, WI, USA. His research interests include air traffic flow management and planning, machine learning, and deep learning-based air traffic management applications.
\end{IEEEbiography}

\begin{IEEEbiography}[{\includegraphics[width=1in,height=1.25in,clip,keepaspectratio]{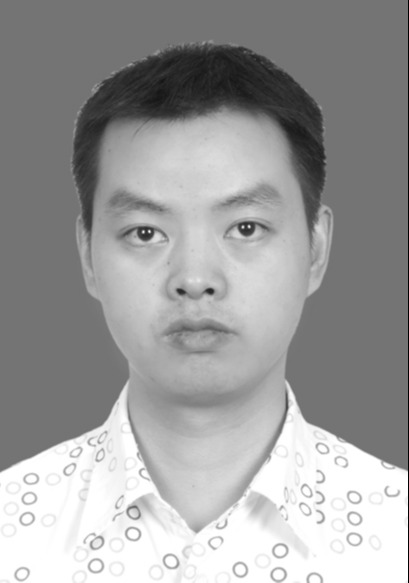}}]
{Hong Liu} received the Ph.D. degree from Sichuan University, Chengdu, China, in 2014. He currently works as an Associate Professor with the College of Computer Science, Sichuan University. His research interests include intelligent transportation systems, and intelligent decision making.
\end{IEEEbiography}





\end{document}